\documentclass[10pt,twocolumn,letterpaper]{article}

\usepackage{cvpr}
\usepackage{times}
\usepackage{epsfig}
\usepackage{graphicx}
\usepackage{amsmath}
\usepackage{amssymb}

\usepackage{multirow,array}
\usepackage[table]{xcolor}
\usepackage{subfigure}
\usepackage[noend]{algorithm2e} %lined,boxed,commentsnumbered
\usepackage{dsfont}
\usepackage{lipsum}
\usepackage{balance} %%%
\usepackage{enumitem}
\usepackage{cite}
\usepackage{appendix}
\graphicspath{ {./images/} }

\usepackage[pagebackref=true,breaklinks=true,letterpaper=true,colorlinks,bookmarks=false]{hyperref}

\cvprfinalcopy % *** Uncomment this line for the final submission

\def\cvprPaperID{1439} % *** Enter the CVPR Paper ID here

\ifcvprfinal\fi
\begin{document}

%%%%%%%%% TITLE
\title{Efficient Diverse Ensemble for Discriminative Co-Tracking}

\author{Kourosh Meshgi, Shigeyuki Oba, Shin Ishii\\
Graduate School of Informatics, Kyoto University\\
606--8501 Yoshida-honmachi, Kyoto, Japan\\
{\tt\small \{meshgi-k,oba,ishii\}@sys.i.kyoto-u.ac.jp}
%\and
}

\maketitle
%\thispagestyle{empty}

%%%%%%%%% ABSTRACT
\begin{abstract}
Ensemble discriminative tracking utilizes a committee of classifiers, to label data samples, which are in turn, used for retraining the tracker to localize the target using the collective knowledge of the committee. Committee members could vary in their features, memory update schemes, or training data, however, it is inevitable to have committee members that excessively agree because of large overlaps in their version space. To remove this redundancy and have an effective ensemble learning, it is critical for the committee to include consistent hypotheses that differ from one-another, covering the version space with minimum overlaps. In this study, we propose an online ensemble tracker that directly generates a diverse committee by generating an efficient set of artificial training. The artificial data is sampled from the empirical distribution of the samples taken from both target and background, whereas the process is governed by query-by-committee to shrink the overlap between classifiers. The experimental results demonstrate that the proposed scheme outperforms conventional ensemble trackers on public benchmarks.
\end{abstract}

%%%%%%%%% BODY TEXT
\section{Introduction}
\begin{figure}[!t]
\subfigure[Typical ensemble state \label{fig:c1}]{\includegraphics[width=0.49\linewidth]{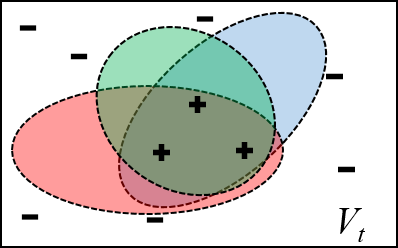}} 
\subfigure[Conventional update \label{fig:c2}]{\includegraphics[width=0.49\linewidth]{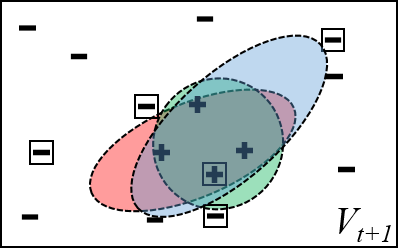}}\\
\subfigure[Partial update \label{fig:c3}]{\includegraphics[width=0.49\linewidth]{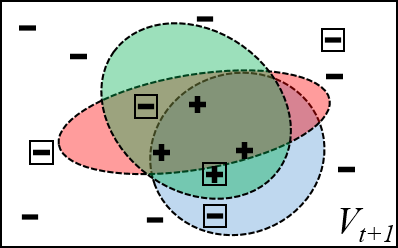}}
\subfigure[Diversified update \label{fig:c4}]{\includegraphics[width=0.49\linewidth]{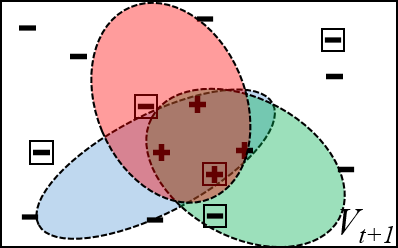}}
\caption{Version space examples for ensemble classifiers. \textbf{(a)} All hypotheses are consistent with the previous labeled data, but each represents a different classifier in the version space $V_t$. In the next time step, the models are updated with the new data (boxed). \textbf{(b)} Updating with all of the data tend to make the hypothesis more overlapping. \textbf{(c)} Random subsets of training data are given to the hypotheses and they update without considering the rest of the data, the hypotheses cover random areas of the version space. \textbf{(d)} Random subsets of training data plus artificial generated data \textit{(proposed)}, trains the hypothese to be mutually uncorrelated as much as possible, while encouraging them to cover more (unexplored) area of the version space. }
\label{fig:diversity}
\vspace{-0.5 cm}
\end{figure}

Tracking-by-detection \cite{grabner2006real,hare2011struck,bai2012robust,grabner2008semi,babenko2009visual,kiani2017learning} as one the most popular approaches of discriminative tracking utilizes classifier(s) to perform the classification task using object detectors. In a tracking-by-detection pipeline, several samples are obtained from each frame of the video sequence, to be classified and labeled by the target detector, and this information is used to re-train the classifier in a closed feedback loop. This approach advantages from the overwhelming maturity of the object detection literature, both in the terms of accuracy and speed \cite{danelljan2016eco,danelljan2016beyond}, yet struggles to keep up with the target evolution as it rises issues such as proper strategy, rate, and extent of the model update \cite{salti2012adaptive,wu2013online,li2016nus}. To adapt to object appearance changes, the tracking-by-detection methods update the decision boundary as opposed to object appearance model in generative trackers. Imperfections of target detection and model update throughout the tracking, manifest themselves as accumulating errors, which essentially drifts the model from the real target distribution, hence leads to target loss and tracking failure.
Such imperfections can be caused by labeling noise, self-learning loop, sensitive online-learning schemes, improper update frequency, non-realistic assumption about the target distribution, and equal weights for all training samples. 

Misclassification of a sample due to drastic target transformations, visual artifacts (such occlusion) or model errors not only degrades target localization accuracy, but also confuses the classifier \cite{hare2011struck} when trained by this erroneous label. Typically in tracking-by-detection, the classifier is retrained using its own output from the earlier tracking episodes (the self-learning loop), which amplitudes a training noise in the classifier and accumulate the error over time. 
The problem amplifies when the tracker lacks a forgetting mechanism or is unable to obtain external scaffolds. Some researchers believe in the necessity of having a ``teacher'' to train the classifier \cite{grabner2008semi}. This inspired the use of co-tracking \cite{tang2007co}, ensemble tracking \cite{saffari2009line,zhang2014meem}, disabling updates during occlusions, or label verification schemes \cite{kalal2012tracking} to break the self-learning loop using auxiliary classifiers.

Ensemble tracking framework provides effective frameworks to tackle one or more of these challenges. In such frameworks, the self-learning loop is broken, and the labeling process is performed by leveraging a group of classifiers with different views \cite{grabner2006real,han2017branchout,saffari2009line}, subsets of training data \cite{meshgi2016robust} or memories \cite{zhang2014meem,meshgi2017active}. 
The main challenge in ensemble methods is how to decorrelate ensemble members and diversify learned models \cite{han2017branchout}.
Combining the outputs of multiple classifiers is only useful if they disagree on some inputs \cite{krogh1995neural}, however, individual learners with similar training data are usually highly correlated \cite{zhou2012ensemble} (Fig. \ref{fig:diversity}).

\textbf{Contributions:} We propose a diversified ensemble discriminative tracker (DEDT) for real-time object tracking. We construct an ensemble using various subsamples of the tracking data and maintain the ensemble throughout the tracking. This is possible by devising methods to update the ensemble to reflect target changes while keeping its diversity to achieve good accuracy and generalization. In addition, breaking the self-learning loop to avoid the potential drift of the ensemble is applied in a co-tracking framework with an auxiliary classifier. However, to avoid unnecessary computation and boost the accuracy of the tracker, an effective data exchange scheme is required. We demonstrate that learning ensembles with randomized subsets of training data along with artificial data with diverse labels in a co-tracking framework achieve superior accuracy. This paper offers the following contributions:

\begin{itemize}[noitemsep]
\item We propose a novel ensemble update scheme that generates necessary samples to diversify the ensemble. Unlike the other model update schemes that ignore the correlation between classifiers of an ensemble, this method is designed to promote diversity.%, see Figure \ref{fig:sample}.
\item We propose a co-tracking framework that accommodates the short and long-term memory mixture, effective collaboration between classification modules, and optimized data exchange between modules by borrowing the concept of query-by-committee \cite{seung1992query} from active learning literature.
\end{itemize} 
In this view, our proposed method is distinguishable from CMT \cite{meshgi2017active} that uses multiple-memory horizons for training the ensemble. It is also different from MUSTer \cite{hong2015multi} that use long-term memory to validate the results of short-memory tracker and TGPR \cite{gao2014transfer}, in which long-term memory regularizes the results of short-memory tracker.
Furthermore, the proposed framework differs from the co-tracking elaborated in \cite{tang2007co}, as in that method two classifiers cast a weighted vote to label the target, and pass the samples they struggle with to the other one to learn. However, in our tracker, the ensemble passes the disputed samples to an auxiliary classifier which is trained on all of the data periodically, to provide the effect of long-term memory while being resistant to abrupt changes, outliers and label noise.
The evaluation results of DEDT on OTB50 \cite{wu2013online}, OTB100\cite{wu2015object}, and VOT2015\cite{kristan2015visual} datasets demonstrates competitive accuracy of DEDT compared to the state-of-the-art of tracking.

\begin{figure}
\includegraphics[width=1\linewidth]{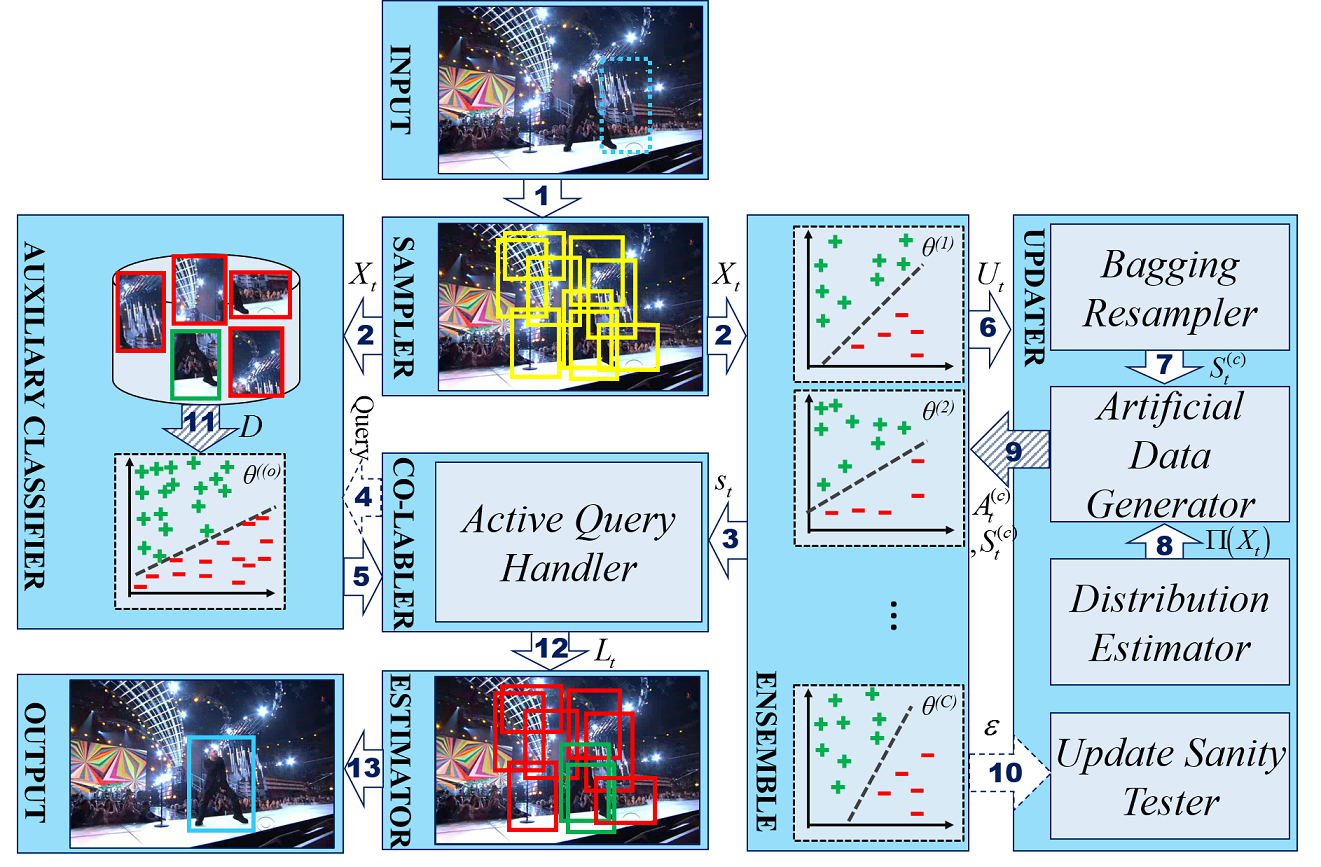}
\caption{Schematic of the system. The proposed tracker, DEDT, labels the obtained sample using an homogeneous ensemble of the classifiers, the committee. The samples that the committee has highest disagreement upon (the uncertain samples) are queried from the auxiliary classifier, a different type of classifier. The location of the target is then estimated using the labeled target. Each member of the ensemble is then updated with a random subset of uncertain samples. By generating the diversity set (r.t. Sec \ref{sec:diversify}), the ensemble is then diversified, yielding a more effective ensemble. For notion and procedure please r.t. Sec \ref{sec:notion} and Alg. \ref{alg:dedt}. }
\label{fig:schematic}
\vspace{-0.5 cm}
\end{figure}

\section{Prior Work}

\noindent\textbf{Ensemble tracking:} Using a linear combination of several weak classifiers with different associated weights has been proposed in a seminal work by Avidan \cite{avidan2007ensemble}. Following this study, constructing an ensemble by boosting \cite{grabner2006real}, online boosting \cite{oza2005online,leistner2010miforests}, multi-class boosting \cite{saffari2010robust} and multi-instance boosting \cite{babenko2009visual,zhang2013real} led to the enhancement of the performance of the ensemble trackers. Despite its popularity, boosting demonstrates low endurance against label noise \cite{santner2010prost} and alternative techniques such as Bayesian ensemble weight adjustment \cite{bai2012robust} has been proposed to alleviate this shortcoming. Recently, ensemble learning based on CNNs gained popularity. Researchers make ensembles of CNNs that shares convolutional layers \cite{nam2016modeling}, different loss functions for each output of the feature map \cite{wang2016stct}, and repeatedly subsampling different nodes and layers in fully connected layers on CNN to build an ensemble \cite{han2017branchout,li2016convolutional}. %
Furthermore, it is proposed to exploit the power of ensembles such as feature adjustment in ensembles \cite{gall2011hough} and the addition of the ensemble's members \cite{saffari2009line,zhang2014meem} over-time.

\noindent\textbf{Ensemble diversity:} Empirically, ensembles tend to yield better results when there is a significant diversity among the models \cite{kuncheva2003measures}. Zhou \cite{zhou2012ensemble} categorizes the diversity generation heuristics into \textit{(i)} manipulation of data samples based on sampling approaches such as bagging and boosting (e.g. in \cite{meshgi2016robust}), \textit{(ii)} manipulation of input features such as online boosting \cite{grabner2006real}, random subspaces\cite{salaheldin2013robust}, random ferns \cite{rao2012online} and random forests \cite{saffari2009line} or combining using different layers, neurons or interconnection layout of CNNs \cite{han2017branchout,li2016convolutional}, \textit{(iii)}
manipulation of learning parameter, and \textit{(iv)} manipulation of the error representation. The literature also suggests a fifth category of manipulation of error function which encourages the diversity such as ensemble classifier selection based on Fisher linear discriminant \cite{visentini2009diversity}.

\noindent\textbf{Training data selection:} A principled ordering of training examples can reduce the cost of labeling and lead to faster increases in the performance of the classifier \cite{vijayanarasimhan2011cost}, therefore we strive to use training examples based on their usefulness, and avoid using on all of them (including noisy ones and outliers) that may result in higher accuracy \cite{de2001robust}. Starting from easiest examples (Curriculum learning) \cite{bengio2009curriculum}, pruning adversarial examples%\footnote{Images with tiny, imperceptible perturbations that fool a classifier into predicting the wrong labels with high confidence \cite{goodfellow2014explaining}.} 
\cite{lu2017safetynet}, excluding misclassified samples from next rounds of training \cite{vezhnevets2007avoiding}, sorting samples by their training value \cite{lapedriza2013all} are some of the proposed approaches in the literature. 
However, the most common setting is active learning, in which the algorithm selects which training examples to label at each step for the highest gains in the performance. In this view, it may require to focus on learning the hardest examples first. For example, following the criteria of ``highest uncertainty'', an active learner select samples closest to the decision boundary to be labeled next. This concept can be useful in visual tracking, e.g. to measure the uncertainty caused by bags of samples \cite{zhang2013robust}.

\noindent\textbf{Active learning for ensembles:}
Query-by-committee (QBC) \cite{seung1992query} is one of the most popular ensemble-based active learning approaches, which constructs a committee of models representing competing hypotheses to label the samples. By defining a utility function on the ensemble (such as disagreement, entropy, or Q-statistics \cite{zhou2012ensemble}), this method selects the most informative samples to be queried from the oracle (or any other collaborating classifier) in a form of the query optimization process \cite{settles2012active}. Built upon randomized component learning algorithm, QBC involves Gibbs sampling, which requires adaptation to use deterministic classifiers. This was realized by resampling different subsets of data to construct an ensemble of deterministic base learners in query-by-bagging and query-by-boosting frameworks \cite{abe1998query}. The set of hypotheses consistent with the data is called \textit{version space} and by selecting the most informative samples to be labeled, QBC attempts to shrink the version space. However, only a committee of hypotheses that effectively samples the version space of all consistent hypotheses is productive for the sample selection \cite{cohn1996active}. To this end, it is crucial to promote the diversity of the ensemble \cite{melville2004diverse}. In QBag and QBoost algorithms, all of the classifiers are trained on random subsets of the similar dataset, which degrade the diversity of the ensemble. Reducing the number of necessary labeled samples \cite{lampert2009active}, unified sample learning and feature selection procedure \cite{li2015active} and reducing the sampling bias by controlling the variance \cite{beygelzimer2009importance} are some of the improvements that active learning provides for the discriminative trackers. Moreover, using diversity data to diversify the committee members \cite{melville2004diverse} and promoting the classifiers that have unique misclassifications \cite{visentini2009diversity} are from few samples that active learning was employed to promote the diversity of the ensemble.

\section{Tracking by Detection}
By definition, a tracker tries to determine the state of the target $\mathbf{p}_t$ in frame $F_t$ ($t \in \{1,\ldots,T\}$) by finding the transformation $\mathbf{y}_t$ from its previous state $\mathbf{p}_{t-1}$. In tracking-by-detection formulation, the tracker employs a classifier $\theta_t$ to separate the target from the background. It is realized by evaluating possible candidates from the expected target state-space $\mathcal{Y}_t$. The candidate whose appearance resembles the target the most, is usually considered as the new target state. Finally, the classifier is updated to reflect the recent information. 

To this end, first several samples $\mathbf{x}_t^{\mathbf{p}_{t-1} \circ \mathbf{y}_t^j} \in \mathcal{X}_t$ are obtained by a transformation $\mathbf{y}^j_t \in \mathcal{Y}_t$ from the previous target state, $\mathbf{p}_{t-1} \circ \mathbf{y}^j_t$. Sample $j \in \{ 1, \ldots , n\}$ indicates the location $\mathbf{p}_{t-1} \circ \mathbf{y}_t^j$ in the frame $F_t$, where the image patch $\mathbf{x}_t^{\mathbf{p}_{t-1} \circ \mathbf{y}_t^j}$ is contained. 
%The transformation space $\mathcal{Y}_t$ is usually defined by motion models \cite{cehovin2013robust}, optical flow \cite{rodriguez2009tracking}, context supports \cite{grabner2010tracking}, confidence maps \cite{tang2007co}, or a combination of these. It also depends on the representation that ranges from 2D translation to 2D affine \cite{kwon2009tracking} or even 3D transformations for holistic or patch-based \cite{adam2006robust} templates. 
%
Then, each sample is evaluated by the classifier scoring function $h: \mathcal{X}_t \rightarrow \mathbb{R}$ to calculate the score 
$
s^j_t = h(\mathbf{x}_t^{\mathbf{p}_{t-1} \circ \mathbf{y}_t^j} | \theta_t)
$.
This score is utilized to obtain a label $\ell^j_t$ for the sample, typically by thresholding its score,
\begin{align}
\ell^j_t &=
  \begin{cases}
   +1        & ,s^j_t > \tau_u \\
   -1        & ,s^j_t < \tau_l \\
   0         & ,\text{otherwise}
  \end{cases}
  \label{eq:label_single}
\end{align}
where $\tau_l$ and $\tau_u$ serves as lower and upper thresholds respectively. 
Finally, the target location $\mathbf{y}_t$ is obtained by comparing the samples' classification scores. To obtain the exact target state, the sample with highest score is selected as the new target, 
$
\mathbf{y}_t = \mathbf{y}_t^{j^*} \; \mathrm{s.t.} \; j^* = \underset{i}{\mathrm{argmax}} \; ( s_t^j ).
$
A subset of the samples and their labels are used to re-train the classifier's model 
$
\theta_{t+1} = u(\theta_t, \mathcal{D}_{\xi(t)}).
$
Here, $\mathcal{D}_t = \{\langle \mathcal{X}_t,\mathcal{L}_t\rangle\}$ is the set of samples $\mathcal{X}_t$ and their labels $\mathcal{L}_t$, $u(.)$ is the model update function, and the $\xi(t)$ defines the subset of the samples that the tracker considers for model update. 
%Many of the adaptive trackers utilize online-learning classifiers \cite{hare2011struck,ross2008incremental} in which only the data from the recent frame ($\xi(t)=\{t\}$) is used. Fixed trackers use only the data from the first frame ($\xi(t) = \{1\}$) and some trackers utilize the samples obtained from several recent frames to update their model ($\xi(t) = \{t-\Delta, \ldots, t-1, t \}$).

An ensemble discriminative tracker employs a set of classifiers instead of one. These classifiers, hereafter called \textit{committee}, are represented by $\mathcal{C}_t=\{\theta_t^{(1)},\ldots,\theta_t^{(C)}\}$, and are typically homogeneous and independent (e.g.,  \cite{saffari2009line,leistner2010miforests}). Popular ensemble trackers utilize the majority voting of the committee as their utility function,
\begin{equation}
s^j_t = \sum_{c=1}^C \mathrm{sign} \big( h(\mathbf{x}_t^{\mathbf{p}_{t-1} \circ \mathbf{y}_t^j} | \theta_t^{(c)}) \big).
\label{eq:score_ensemble}
\end{equation}
Then eq\eqref{eq:label_single} is used to label the samples. 

The model of each classifier is updated independently,
$
\theta^{(c)}_{t+1} = u(\theta^{(c)}_t, \mathcal{D}_{\xi(t)})
$
meaning that all of the committee members are trained with a similar set of samples and a common label for them. 

\section{Diverse Ensemble Discriminative Tracker}
\label{sec:dedt}
We propose a diverse ensemble tracker composed of a highly-adaptive and diverse ensemble of classifiers $\mathcal{C}$ (the \textit{committee}), a long-term memory object detector (that serves as the \textit{auxiliary} classifier), and an information exchange channel governed by active learning. This allows for effective diversification of the ensemble, improving the generalization of the tracker and accelerating its convergence to the ever-changing distribution of target appearance. We leveraged the complementary nature and long-term memory of the auxiliary tracker to facilitate effective model update.

One way to diversify the ensemble is to increase the number of examples they disagree upon \cite{krogh1995neural}. Using bagging and boosting to construct an ensemble out of a fix sample set, ignores this critical need for diversity as all of the data are randomly sampled from a shared data distribution. However, for each committee member, there exists a set of samples that distinguish them from other committee members. One way to obtain such samples is to generate some training samples artificially to differ maximally from the current ensemble \cite{melville2003constructing}. 

The diversified ensemble covers larger areas of the version space (i.e. the space of consistent hypotheses with the samples from current frame), however, this radical update of the ensemble may render the classifier susceptible to drastic target appearance changes, abrupt motion, and occlusions. In this case, given the non-stationary nature of the target distribution\footnote{The non-stationarity means that the appearance of an object may change so significantly that a negative sample in the current frame looks more similar to a positive example in the previous frames~\cite{bai2013randomized}.
}, the classifier should adapt itself rapidly with the target changes, yet it should keep a memory of the target to re-identify if the target goes out-of-view or got occluded (as known as stability-plasticity dilemma \cite{grabner2008semi}). In addition, there are samples for which the ensemble is not unanimous and an external teacher maybe deemed required.

To amend these shortcomings, an auxiliary classifier is utilized to label the samples which the ensemble dispute upon (co-tracking). This classifier is batch-updated with all of the samples less frequently than the ensemble, realizing the longer memory for the tracker. Active query optimization is employed to query the label of the most informative samples from the auxiliary classifier, which is observed to effectively balance the stability-plasticity equilibrium of the tracker as well. Figure \ref{fig:schematic} presents the schematic of the proposed tracker.

\subsection{Formalization}
\label{sec:notion}
In this approach, if the committee comes to a solid vote about a sample, then the sample is labeled accordingly. However, when the committee disagrees about a sample, its label is queried from the auxiliary classifier $\theta^{(o)}_t$:
\begin{align}
\ell^j_t &=
  \begin{cases}
   +1        & ,s^j_t > \tau_u \\
   -1        & ,s^j_t < \tau_l \\
    \mathrm{sign} \big( h(\mathbf{x}_t^{\mathbf{p}_{t-1} \circ \mathbf{y}_t^j} | \theta^{(o)}_t) \big)         & ,\text{otherwise}
  \end{cases}
  \label{eq:label_dedt}
\end{align}
in which $s^j_t$ is derived from eq\eqref{eq:score_ensemble}. The uncertain samples list is defined as $\mathcal{U}_t = \{ \mathbf{x}_t^{\mathbf{p}_{t-1} \circ \mathbf{y}_t^j} | \tau_l < s^j_t < \tau_u  \}$.

The committee members are then updates using our proposed mechanism $f(.)$ using the uncertain samples $\mathcal{U}_t$, 
\begin{equation}
\theta^{(c)}_{t+1} = f(\theta^{(1..c)}_t, \mathcal{U}_t, \mathcal{D}_t)
\end{equation}
Finally, to maintain a long-term memory and slower update rate for the auxiliary classifier, it is updated every $\Delta$ frames with all of the samples from $t-\Delta$ to $t$.
\begin{align}
\theta^{(o)}_{t+1} = 
    \begin{cases}
    u(\theta^{(o)}_t, \mathcal{D}_{t-\Delta..t})     &, \mathrm{if} \; t \neq k\Delta + 1 \\
    \theta^{(o)}_{t}                                             &, \mathrm{if} \; t = k\Delta + 1
    \end{cases}
\label{eq:update_dedt}
\end{align} 
Algorithm \eqref{alg:dedt} summarizes the proposed tracker.

%%%%%%%%%%%%%%
\begin{algorithm}[h]
\DontPrintSemicolon
\SetKwInOut{Input}{input}\SetKwInOut{Output}{output}
\SetKwRepeat{Do}{do}{while}
\Input{Committee models $\theta_{t}^{(c)}$, Auxiliary model $\theta^{(o)}$}
\Input{Target position in previous frame $\mathbf{p}_{t-1}$}
\Output{Target position in current frame $\mathbf{p}_{t}$}
\BlankLine
\For{$j \leftarrow 1$ \KwTo $n$ }
{
\emph{Sample a transformation $\mathbf{y}_t^j \sim \mathcal{N}(\mathbf{p}_t,\Sigma_{search})$}\;
\emph{Calculate committee score $s_t^j$} (eq\eqref{eq:score_ensemble})\;
\uIf(sample label is uncertain){$\tau_l < s_t^j < \tau_u$}
        {
        \emph{$\ell_t^j = \mathrm{sign} \big( h(\mathbf{x}_t^{\mathbf{p}_{t-1} \circ \mathbf{y}_t^j} | \theta^{(o)}) \big)$}\;
        \emph{$\mathcal{U}_t \leftarrow \mathcal{U}_t \cup \{\langle \mathbf{x}^{\mathbf{p}_{t-1} \circ \mathbf{y}^j_t} , \ell^j_t \rangle\}$}\;
        }
\Else
        {
        \emph{$\ell_t^j = \mathrm{sign} (s_t^j)$}\;
        }
\emph{$\mathcal{D} \leftarrow \mathcal{D} \cup \{\langle \mathbf{x}^{\mathbf{p}_{t-1} \circ \mathbf{y}^j_t} , \ell^j_t \rangle\}$}\;
}% endloop sample 
\For{$c \leftarrow 1$ \KwTo $C$ }
{
\emph{Uniformly resample $m$ data $\mathcal{S}_{t}^{(c)}$ from $\mathcal{U}_t$}\;
\emph{ $\theta'^{(c)}_t \leftarrow u(\theta_t^{(c)}|\mathcal{S}_{t}^{(c)})$}\;
}
\emph{Calculate the prediction error of $\mathcal{C}_t$, $\epsilon(\mathcal{C}_t)= \frac{|\mathcal{U}_t|}{|\mathcal{D}_t|}$}\;
\emph{Calculate empirical distribution of samples, $\Pi(\mathcal{X}_t)$}\;

\For{$c \leftarrow 1$ \KwTo $C$ }
{
\Do{$\epsilon(\mathcal{C}''_t) \geq \epsilon (\mathcal{C}_t)$}
{
\emph{Draw $m'$ samples $\mathcal{A}_{t}^{(c)}$ from $\Pi(\mathcal{X}_t)$}\;
\emph{Calculate class membership probability $\hat{\ell}(\mathcal{C}'_t)$}\;
\emph{Set the labels of samples $\propto \frac{1}{\hat{\ell}(\mathcal{C}'_t)}$}\;
\emph{ $\theta''^{(c)}_t \leftarrow u(\theta'^{(c)}_t|\mathcal{A}_{t}^{(c)})$}\;
\emph{Calculate new prediction error $\epsilon (\mathcal{C}''_t)$} (eq\eqref{eq:update_pred_error})\;
} % end do-while
$\theta'^{(c)}_{t} \leftarrow \theta''^{(c)}_t$
}
\emph{All diversity sets are applied, $\mathcal{C}_{t+1} \leftarrow \mathcal{C}'_t$}\;
\If{$\mathrm{mod}(t,\Delta) = 0$}
{
\emph{$\theta^{(o)}_{t+1} \leftarrow u(\theta^{(o)}_t, \mathcal{D}_{t-\Delta..t})$}\;
}% end if
\emph{Target transformation $\mathbf{y}_t = \mathbf{y}_t^{j^*} \mathrm{s.t.}  j^* = \underset{i}{\mathrm{argmax}}  (s_t^j)$} \;
\emph{Calculate target position $\mathbf{p}_t = \mathbf{p}_{t-1} \circ \mathbf{\hat{y}}_t$}\;
\BlankLine
\caption{Diverse Ensemble Discriminative Tracker}
\label{alg:dedt}
%\vspace{-0.5cm}
\end{algorithm}
%%%%%%%%%%%%%%
%

\subsection{Diversifying Ensemble Update}
\label{sec:diversify}
The model updates to construct a diverse ensemble either replace the weakest or oldest classifier of the ensemble \cite{grabner2006real,avidan2007ensemble} or creates a new ensemble in each iteration \cite{melville2004diverse}. While the former lacks flexibility to adjust to the rate of target change, the latter involves a high level of computation redundancy. To alleviate these shortcomings, we create an ensemble for the first frame, update them in each frame to keep a memory of the target, and diversify them to improve the effectiveness of ensemble.
The diversifying update procedure is as follows: 
\begin{enumerate}[noitemsep]
\item The members ensemble $\mathcal{C}_t$ is updated with a random subsets (of size $m$) of the uncertain data $\mathcal{U}_t$, that make them more adept in handling such samples, and generate a temporary ensemble $\mathcal{C}'_t$. Note that for certain samples (those not in $\mathcal{U}_t$), the committee is unanimous about the label and adding them to the training set of the committee classifiers is redundant \cite{meshgi2016robust}.
\item The label prediction of the original ensemble $\mathcal{C}_t$ is then calculated on $\mathcal{D}_t$ w.r.t. the labels given by the whole tracker (composed of the ensemble and the auxiliary classifier), and prediction error $\epsilon(\mathcal{C}_t)$ is obtained.
\item The empirical distribution of training data, $\Pi(\mathcal{X}_t)$, is calculated to govern the creation of the artificial data. 
\item In an iterative process for each of the committee members, $m'$ samples are drawn from a $\Pi(\mathcal{X}_t)$, assuming attribute independence. Given a sample, the class membership probabilities of the temporary ensemble $\hat{\ell}(\mathcal{C}'_t)$ that is the probability of selecting a label by the temporary ensemble on $\mathcal{D}_t$, is then calculated. Labels are then sampled from this distribution, such that the probability of selecting a label is inversely proportional to the temporary ensemble prediction. This set of artificial samples and their diverse labels are called the \textit{diversity set} of committee member $c$, $\mathcal{A}^{(c)}_t$.
\item The classifier $c$ of temporary ensemble is updated with $\mathcal{A}^{(c)}_t$, to obtain the diverse ensemble $\mathcal{C}''_t=\{\theta''^{(c)}_t\}$ and calculate its prediction error $\epsilon(\mathcal{C}''_t)$. If this update increases the total prediction error of the ensemble ($\epsilon(\mathcal{C}''_t) > \epsilon (\mathcal{C}_t)$), then the artificial data is rejected and new data $\mathcal{A}^{(c)}_t$ should be generated,
\begin{equation}
\epsilon(\mathcal{C}''_t) = \sum_{c=1}^C \sum_{j=1}^n \mathds{1} \big( \ell^j_t \ne h(\mathbf{x}_t^{\mathbf{p}_{t-1} \circ \mathbf{y}_t^j} | \theta''^{(c)}_t) \big).
\label{eq:update_pred_error}
\end{equation}
\end{enumerate}
where $\mathds{1}(.)$ denotes the step function that returns 1 iff its argument is true/positive and 0 otherwise.

This procedure creates samples for each member of the committee that distinguish them from other members of the ensemble using a contradictory label (therefore improving the ensemble diversity \cite{melville2004diverse}), but only accepts them when using such artificial data improves the ensemble accuracy.

\subsection{Implementation Details}
There are several parameters in the system such as the number of committee members ($C$), parameters of sampling step (number of samples $n$, effective search radius $\Sigma_{search}$), and the holding time of auxiliary classifier ($\Delta$). Larger values of $m$ results in temporary committee with a higher degree of overlap, thus less diverse, whereas smaller values of $m$ tend to miss the latest changes of the quick-changing target. A Larger number of artificial samples $m'$ result in more diversity in the ensemble, but reduce the chance of successful update (i.e. lowering the prediction error of the ensemble). These parameters were tuned using a simulated annealing optimization on a cross-validation set. 

In our implementation, we used kd-tree-based KNN classifiers with HOG \cite{dalal2005histograms} feature for the ensemble and reused the calculations with a caching mechanism to accelerate classification. 
For the empirical distribution of the data, a Gaussian distribution is determined by estimating the mean and standard variation of the given training set (i.e. HOG of $\mathcal{X}_t$). In addition, to localize the target, the samples with the highest sum of confidence scores is selected as the next target position.
The auxiliary classifier is a a part-based detector \cite{felzenszwalb2010object}. The features, part-base detector dictionary, and the parameters of committee members ($k$ of KNNs), thresholds $\tau_l,\tau_u$, and the rest of above-mentioned parameters (Except for $C$ that have been adjusted to control the speed of the tracker, here) have been adjusted using cross-validation. With $C=15, k=23, n=1000, m=80, m'= 250, \tau_u=0.54$ and $\tau_l=-0.41$ DEDT achieved the speed of 21.97 fps on a Pentium IV PC @ 3.5 GHz and a Matlab/C++ implementation on a CPU. Source code can be found at \url{http://ishiilab.jp/member/meshgi-k/dedt.html}.

\section{Experiments}
%%%%%%%%
\begin{table}[t]
\caption{Quantitative evaluation of trackers under different visual tracking challenges of OTB50 \cite{wu2013online} using AUC of success plot and their overall precision. The {\color{red}first}, {\color{green}second} and {\color{blue}third} best methods are shown in color. More data are available on  \url{http://ishiilab.jp/member/meshgi-k/dedt.html}.}
\label{tab:attributes}
\centering
\scalebox{0.72}{
\renewcommand{\arraystretch}{1.1}
\begin{tabular}{@{}l@{} @{}c @{}c @{}c @{}c @{}c @{}c @{}c @{}c @{}c @{}c@{}}
\hline
\makebox[10mm]{Attribute} & \makebox[10mm]{\small TLD} & \makebox[10mm]{\small STRK} & \makebox[10mm]{\small TGPR} & \makebox[10mm]{\small MEEM}& \makebox[10mm]{\small MSTR} & \makebox[10mm]{\small STPL} & \makebox[10mm]{\small CMT} &\makebox[10mm]{\small SRDCF}& \makebox[10mm]{\small CCOT}& \makebox[10mm]{\small Ours} \\ \hline
IV     & 0.48 & 0.53 & 0.54 & 0.62 & {\color{blue}0.73} & 0.68 & {\color{blue}0.73} & 0.70 & {\color{red}0.75} & {\color{red}0.75} \\
DEF    & 0.38 & 0.51 & 0.61 & 0.62 & {\color{green}0.69} & {\color{red}0.70} & {\color{green}0.69} & 0.67 & {\color{green}0.69} & {\color{green}0.69} \\
OCC    & 0.46 & 0.50 & 0.51 & 0.61 & 0.69 & 0.69 & 0.69 & {\color{blue}0.70} & {\color{red}0.76} & {\color{green}0.72} \\
SV     & 0.49 & 0.51 & 0.50 & 0.58 & 0.71 & 0.68 & {\color{blue}0.72} & 0.71 & {\color{red}0.76} & {\color{green}0.74} \\
IPR    & 0.50 & 0.54 & 0.56 & 0.58 & 0.69 & 0.69 & {\color{red}0.74} & 0.70 & {\color{blue}0.72} & {\color{green}0.73} \\
OPR    & 0.48 & 0.53 & 0.54 & 0.62 & 0.70 & 0.67 &  {\color{blue}0.73} & 0.69 & {\color{red}0.74} & {\color{red}0.74} \\
OV     & 0.54 & 0.52 & 0.44 & 0.68 & {\color{blue}0.73} & 0.62 & 0.71 & 0.66 & {\color{red}0.79} & {\color{green}0.76} \\
LR     & 0.36 & 0.33 & 0.38 & 0.43 & 0.50 & 0.47 & 0.55 & {\color{red}0.58} & {\color{red}0.70} & {\color{green}0.58} \\
BC     & 0.39 & 0.52 & 0.57 & 0.67 & {\color{green}0.72} & 0.67 & 0.69 & {\color{blue}0.70} & {\color{blue}0.70} & {\color{red}0.73} \\
FM     & 0.45 & 0.52 & 0.46 & 0.65 & 0.65 & 0.56 & {\color{blue}0.70} & 0.63 & {\color{green}0.72} & {\color{red}0.74} \\
MB     & 0.41 & 0.47 & 0.44 & 0.63 & 0.65 & 0.61 & 0.65 & {\color{blue}0.69} & {\color{red}0.72} & {\color{red}0.72} \\
\hline
\hline
Avg. Succ    & 0.49 & 0.55 & 0.56 & 0.62 & 0.72 & 0.69 & {\color{blue}0.72} & 0.70 & {\color{red}0.75} & {\color{green}0.74} \\
Avg. Prec 	 & 0.60 & 0.66 & 0.68 & 0.74 & 0.82 & 0.76 & {\color{blue}0.83} & 0.78 & {\color{red}0.84} & {\color{red}0.84} \\ 
$IoU>0.5$    & 0.59 & 0.64 & 0.66 & 0.75 & 0.86 & 0.82 & {\color{blue}0.83} & {\color{blue}0.83} & {\color{red}0.90} & {\color{green}0.89} \\
\hline
Avg FPS      & 21.2 & 11.3 & 3.7 & 14.2 & 8.3 & {\color{red}48.1} & {\color{green}21.9} & 4.3 & 0.2 & {\color{green}21.9} \\
\hline
\end{tabular}
}
%\vspace{-1.2em}
\end{table}

For our component analysis, we used the OTB50 \cite{wu2013online} dataset and its subsets with a distinguishing attribute to evaluate the tracker performance. These attributes are illumination variation (\textit{IV}), scale variation (\textit{SV}), occlusions (\textit{OCC}), deformation (\textit{DEF}), motion blur (\textit{MB}), fast motion (\textit{FM}), in-plane-rotation (\textit{IPR}), out-of-plane rotation (\textit{OPR}), out-of-view (\textit{OV}), low resolution (\textit{LR}), and background clutter (\textit{BC}), defined based on the biggest challenges that a tracker may face throughout tracking. Additionally, to compare our proposed algorithm against the state-of-the-art we employed OTB100 \cite{wu2015object} and VOT2015 \cite{kristan2015visual} datasets.

For this comparison, we have used success and precision plots, where their area under curve provides a robust metric for comparing tracker performances \cite{wu2013online}. 
The precision plot compares the number of frames that a tracker has certain pixels of displacement, whereas the overall performance of the tracker is measured by the area under the surface of its success plot, where the success of tracker in time $t$ is determined when the normalized overlap of the tracker target estimation $\mathbf{p}_t$ with the ground truth $\mathbf{p}_t^*$ (also known as IoU) exceeds a threshold $\tau_{ov}$. Success plot, graphs the success of the tracker against different values of the threshold $\tau_{ov}$ and its $AUC$ is calculated as
\begin{equation}
AUC = \frac{1}{T} \int_0^1 \sum_{t=1}^T \mathds{1} \left( \frac{ | \mathbf{p}_t \cap \mathbf{p}_t^* | }{ | \mathbf{p}_t^* \cup \mathbf{p}_t^* | } > \tau_{ov} \right) d_{\tau_{ov}},
\label{eq:auc}
\end{equation}
where $T$ is the length of sequence, $|.|$ denotes the area of the region and $\cap$ and $\cup$ stands for intersection and union of the regions respectively. We also compare all the trackers by the success rate at the conventional thresholds of 0.50 ($IoU > 0.50$) \cite{wu2013online}.
The result of the algorithms are reported as the average of five independent runs.

\subsection{Effect of Diversification}
To demonstrate the effectiveness of the proposed diversification method we compare the DEDT tracker with two different versions of the tracker. In the firs version, DEDT-bag, the ensemble classifiers are only updated with uniform-picked subsets of the uncertain data (step 1 in section \ref{sec:diversify}). In the other version, DEDT-art, the committee members are only updated with artificially generated data (steps 2-5 in the same section). All three algorithms use $m+m'$ samples to update their classifiers. 
In addition to the overall performance of the tracker, we measure the diversity of the ensemble using the Q-statistics as elaborated in \cite{kuncheva2003measures}. For statistically independent classifiers $i$ and $j$, the expectation of $Q_{i,k} = 0$. Classifiers that tend to classify the same sample correctly will have positive values of $Q$, and those which commit errors on different samples have negative $Q$ ($-1 \le Q_{i,k} \le +1$). For the ensemble of $C$ classifiers, the averaged Q statistics over all pairs of classifiers is
\begin{align}
Q_{av} &= \frac{2}{C(C-1)}\sum_{i=1}^{C-1} \sum_{j=i+1}^C Q_{i,j} \;,\; \mathrm{s.t.} \\
Q_{i,j} &= \frac{N^{ff}N^{bb}-N^{fb}N^{bf}}{N^{ff}N^{bb}+N^{fb}N^{bf}}
\label{eq:qstat}
\end{align}
where $N^{fb}$ is the number of cases that classifier $i$ classified the sample as foreground, while classifier $j$ detected it as background, etc.

%%%%%%%%
\begin{figure*}[!t]
  \begin{center}
  \subfigure[The diversification procedure\label{fig:decorate-bag}]{\includegraphics[width=0.32\linewidth]{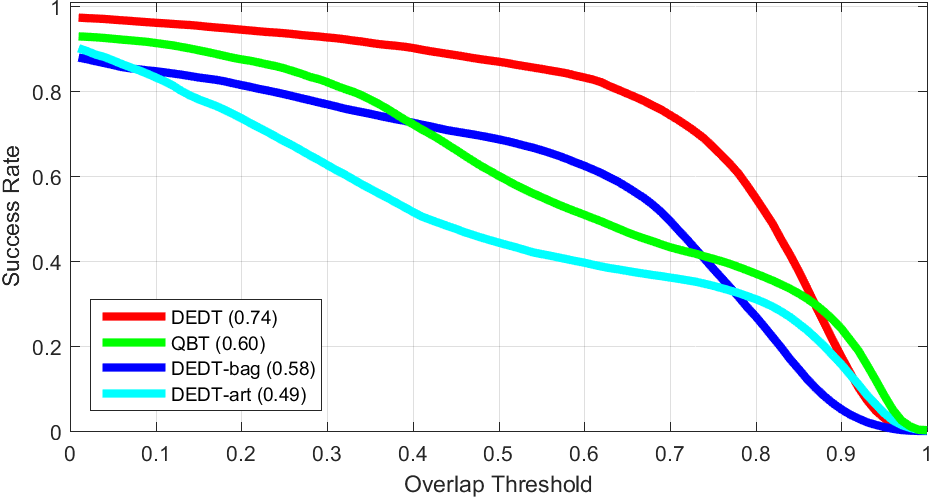}}
  \subfigure[Using artificial data versus real data\label{fig:art_real}]{\includegraphics[width=0.32\linewidth]{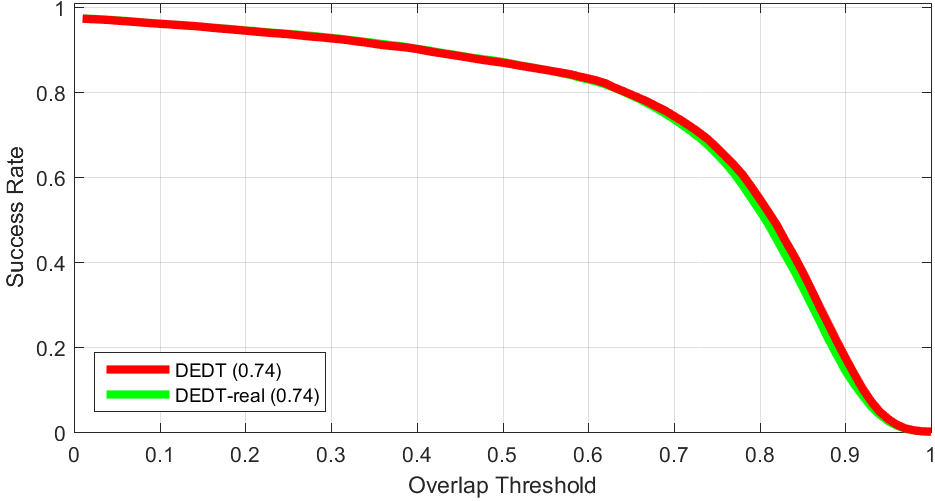}} 
  %\subfigure[Using long-term memory for aux. classifier\label{fig:long_term}]{\includegraphics[width=0.32\linewidth]{success_plot_long}}
  \subfigure[The ``activeness'', i.e. the effect of thresholds\label{fig:delta}]{\includegraphics[width=0.32\linewidth]{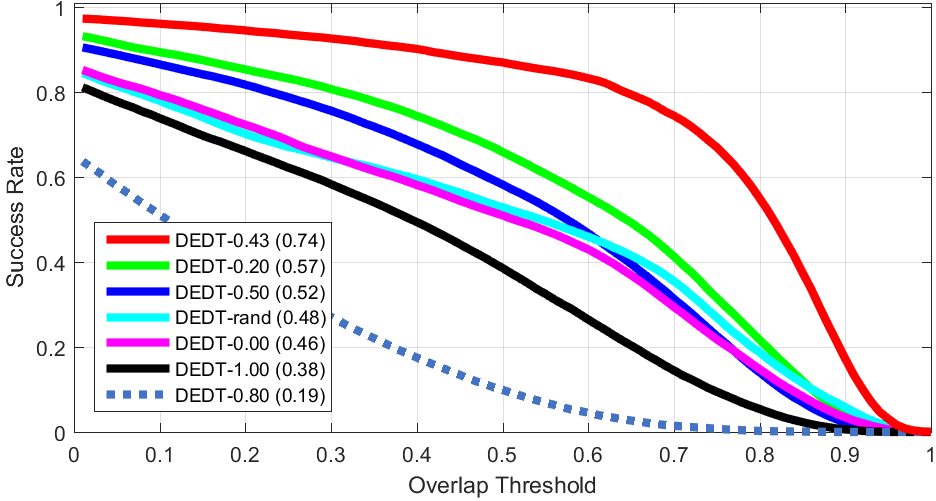}} 
   \end{center}
  \caption{The effect of different components of the proposed algorithm on the overall tracking results on OTB50 \cite{wu2013online}.}
\vspace{-0.5cm}
\end{figure*}

Figure \ref{fig:decorate-bag} illustrates the effectiveness of the diversification mechanism in contrast with merely generating  data or update the classifiers with uninformed subsamples of the data. From the experiment results, $AUC(\text{DEDT-art}) < AUC(\text{DEDT-bag}) < AUC(\text{DEDT})$ and $0 < Q_{av}(\text{DEDT}) < Q_{av}(\text{DEDT-art}) < Q_{av}(\text{DEDT-bag})$ it can be concluded that all of steps of proposed diversification are crucial to maintain an accurate and diverse ensemble. $Q_{av}(\text{DEDT-art}) < Q_{av}(\text{DEDT-bag})$ shows that the diversity of DEDT-art is better than random diversity obtain by DEDT-bag, however, $AUC(\text{DEDT-art}) < AUC(\text{DEDT-bag})$ reveals that merely using artificial data without the samples gathered by the tracker, does not provide enough data for an accurate model update.

\subsection{Effect of using Artificial Data}
In the first look, using synthesized data to train the ensemble that will keep track of a real object may not seem proper. In this experiment, we look for the closest patch of the real image (frame $t$ of the video) to the synthesized sample, and use it as the diversity data. To this end, in each frame, a dense sampling over the frame is performed, the HOG of these image patches are calculated, and the closest match to the generated sample (using Euclidean distance) is selected. The obtained tracker is referred as DEDT-real, and its performance is compared to the original DEDT.

As Figure \ref{fig:art_real} shows, the use of this computationally-expensive version of the algorithm does not improve the performance significantly. However, it should be noted that generating adversarial samples of the ensemble \cite{goodfellow2014explaining} for as the diversity data of individual committee members is expected to increase the accuracy of the ensemble, yet it is out of the scope of the current research and may be considered as a future direction for this research.

%\subsection{Effect of Combining Long and Short Memories}
%Researchers have been combining long-term and short-term classifiers to realize robust tracking. TGPR categorized samples into auxiliary samples from early frames and update them slowly and carefully, and target samples from recent samples that are updated quickly and aggressively \cite{gao2014transfer}. MEEM selects an snapshot of the classifier trained by the samples obtained from the beginning of the tracking to role-back inappropriate updates of the classifier \cite{zhang2014meem}. MUSTer archives consistent key-points of the target in he long-term memory, and validates the tracking of the short-term tracker \cite{hong2015multi}. In our proposed tracker, however, an ensemble of short-memory classifiers invoke the long-term memory when deemed necessary and an active query mechanism governs this process to balance the use of long and short term memories. To see the performance of this scheme, we made DEDT-first that trains the auxiliary classifier on the first frame and do not update this classifier, DEDT-short that updates the auxiliary classifier on each frame, canceling its long-memory properties, and DEDT-isolated that isolate the ensemble from auxiliary classifier, and fuse their results in the end similar to \cite{hong2015multi}. Figure \ref{fig:long_term} shows that both of such strategies have inferior performance in our settings, which promotes the role of active query selection and loopy update of the auxiliary classifier.

\subsection{Effect of ``Activeness''}
%%%%%%%%%
%\begin{figure}[h]
%  \begin{center}
%    \includegraphics[width=0.7\linewidth]{success_plot_delta}
%  \end{center}
%  \caption{The ``activeness'', i.e. the effect of labeling thresholds on the performance of the proposed algorithm on OTB50 \cite{wu2013online}.}
%  \label{fig:delta}
%%  \vspace{-1.8em}
%\end{figure}
Labeling thresholds ($\tau_l$ and $\tau_u$) control the ``activeness'' of the data exchange between the committee and the auxiliary classifier, therefore allowing the ensemble to get more/less assistance for its collaborator. In our implementation, these two values are treated independently, but for the sake of argument assume that $\tau_l = -\delta$ and $\tau_u = +\delta$ ($\delta \in [0,1]$). Figure \ref{fig:delta} compares the effects of different values of the $\delta$, and also a ``random'' data exchange scheme in which the labeler gets the label of the sample from the ensemble or auxiliary classifier with the same chance. To interpret this figure it is prudent to note that $\delta \rightarrow 0$ forces the ensemble to label all of the samples without any assistance from the auxiliary classifier. By increasing $\delta$ the ensemble starts to query highly disputed samples from the auxiliary classifier, which is desired by design. If this value increases excessively, the ensemble queries even slightly uncertain samples from the auxiliary classifier, rendering the tracker prone to the labeling noise of this classifier. In addition, the tracker loses its ability to update rapidly in the case of an abrupt change in the target's appearance or location, leading to a degraded performance of the tracker. In the extreme case of $\delta \rightarrow 1$ the tracker reduces to a single object detector modeled by the auxiliary classifier. 

The information exchange in one way is in the form of querying the most informative labels from the auxiliary classifier, and on the other way is re-training it with the labeled samples by the committee (for certain samples). We observed that this exchange is essential to construct a robust and accurate tracker. Moreover, such data exchange not only breaks the self-learning loop but also manages the plasticity-stability equilibrium of the tracker. In this view, lower values of $\delta$ correspond to a more-flexible tracker, while higher values make it more conservative.

\subsection{Comparison with State-of-the-Art}
To establish a fair comparison with the state-of-the-art, some of the most successful popular discriminative trackers (according to a recent large benchmark \cite{wu2013online,kristan2015visual,wu2015object} and the recent literature) are selected: TLD \cite{kalal2012tracking}, STRK \cite{hare2011struck}, TGPR \cite{gao2014transfer}, MEEM \cite{zhang2014meem}, MUSTer \cite{hong2015multi}, STAPLE \cite{bertinetto2016staple}, CMT \cite{meshgi2017active}, SRDCF \cite{danelljan2015learning}, and CCOT \cite{danelljan2016beyond}.
%%%%%%%%
\begin{figure}[h]
  \begin{center}
    \includegraphics[width=1\linewidth]{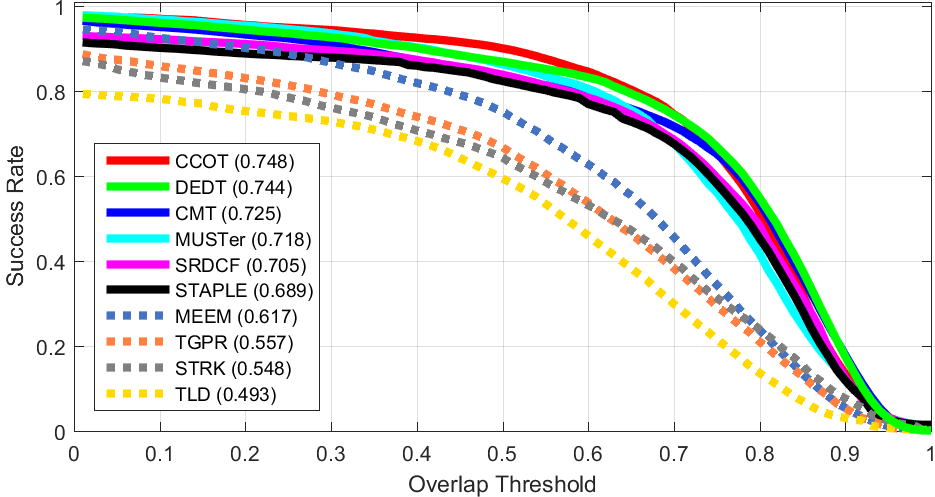}
    \includegraphics[width=1\linewidth]{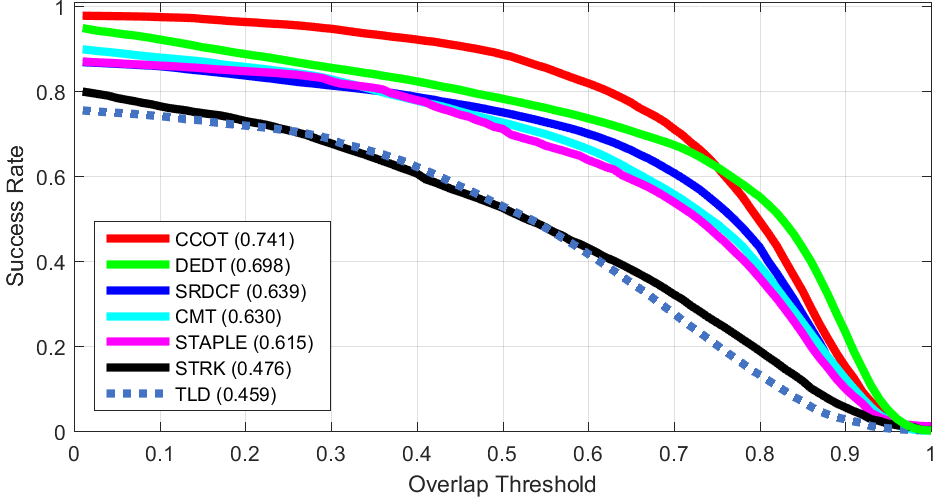}
  \end{center}
  \caption{Quantitative performance comparison of the proposed tracker, DEDT, with the state-of-the-art trackers using success plot on OTB50 \cite{wu2013online} (\textbf{top}) and OTB100 \cite{wu2015object} (\textbf{bottom}).}
  \label{fig:success}
%  \vspace{-1.8em}
\end{figure}
%%%%%%%%

%
%%%%%%%%%
\begin{table}[t]
\caption{Quantitative evaluation of trackers under different visual tracking challenges of OTB100 \cite{wu2015object}.}
\label{tab:otb100}
\centering
\scalebox{0.73}{
\renewcommand{\arraystretch}{1.1}
\begin{tabular}{@{}l@{} @{}c @{}c @{}c @{}c @{}c @{}c @{}c @{}c @{}c @{}c@{}}
\hline
\makebox[10mm]{} & \makebox[10mm]{\small TLD} & \makebox[10mm]{\small STRK} & \makebox[10mm]{\small MEEM}& \makebox[10mm]{\small STPL} & \makebox[10mm]{\small CMT} &\makebox[10mm]{\small SRDCF}& \makebox[10mm]{\small CCOT}& \makebox[10mm]{\small Ours} \\ \hline
Avg. Succ    & 0.46 & 0.48 & {\color{blue}0.65} & 0.62 & 0.63 & 0.64 & {\color{red}0.74} & {\color{green}0.69} \\
Avg. Prec 	 & 0.58 & 0.59 & 0.62 & 0.73 & {\color{blue}0.74} & 0.71 & {\color{red}0.85} & {\color{green}0.81} \\
$IoU>0.5$    & 0.52 & 0.52 & 0.62 & 0.71 & 0.72 & {\color{blue}0.75} & {\color{red}0.88} & {\color{green}0.78} \\
\hline

\end{tabular}
}
%\vspace{-1.2em}
\end{table}

%%%%%%%%%

\begin{table}[t]
\caption{Evaluation on VOT2015 \cite{kristan2015visual} by the means of robustness and accuracy.}
\label{tab:vot}
\centering
\scalebox{0.73}{
\renewcommand{\arraystretch}{1.1}
\begin{tabular}{@{}l@{} @{}c @{}c @{}c @{}c @{}c @{}c @{}c @{}c @{}c@{}}
\hline
\makebox[10mm]{} & \makebox[10mm]{\small STRK} & \makebox[10mm]{\small TGPR} & \makebox[10mm]{\small MEEM}& \makebox[10mm]{\small MSTR} & \makebox[10mm]{\small STPL} & \makebox[10mm]{\small CMT} &\makebox[10mm]{\small SRDCF}& \makebox[10mm]{\small CCOT}& \makebox[10mm]{\small Ours} \\ \hline
Accuracy     & 0.47 & 0.48 & 0.50 & 0.52 & 0.53 & 0.49 & {\color{green}0.56} & {\color{blue}0.54} & {\color{red}0.58} \\
Robustness   & {\color{blue}1.26} & 2.31 & 1.85 & 2.00 & 1.35 & 1.81 & {\color{green}1.24} & {\color{red}0.82} & 1.36 \\
\hline
\end{tabular}
}
%\vspace{-1.2em}
\end{table}
%%%%%%%%%

\begin{figure}[t]
\begin{center}
\includegraphics[width= 0.32\linewidth]{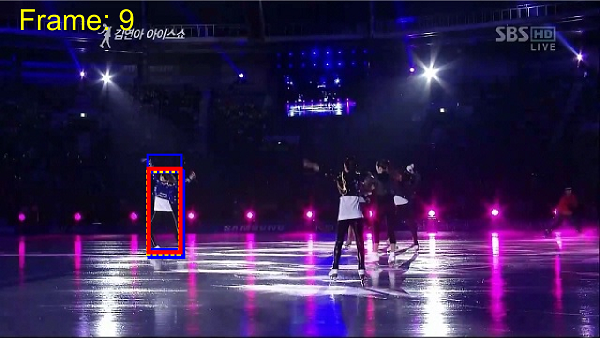}
\includegraphics[width= 0.32\linewidth]{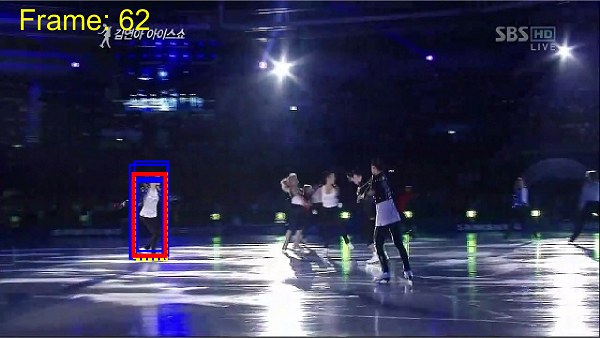}
\includegraphics[width= 0.32\linewidth]{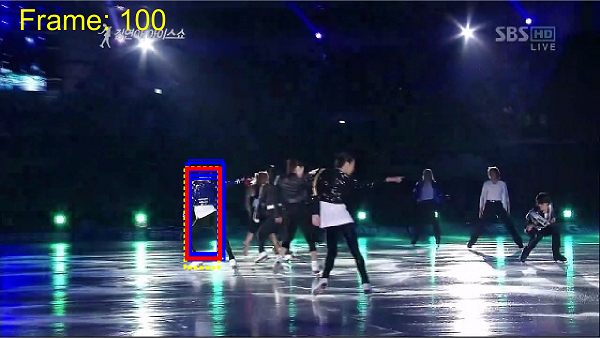}
\\
\includegraphics[width= 0.32\linewidth]{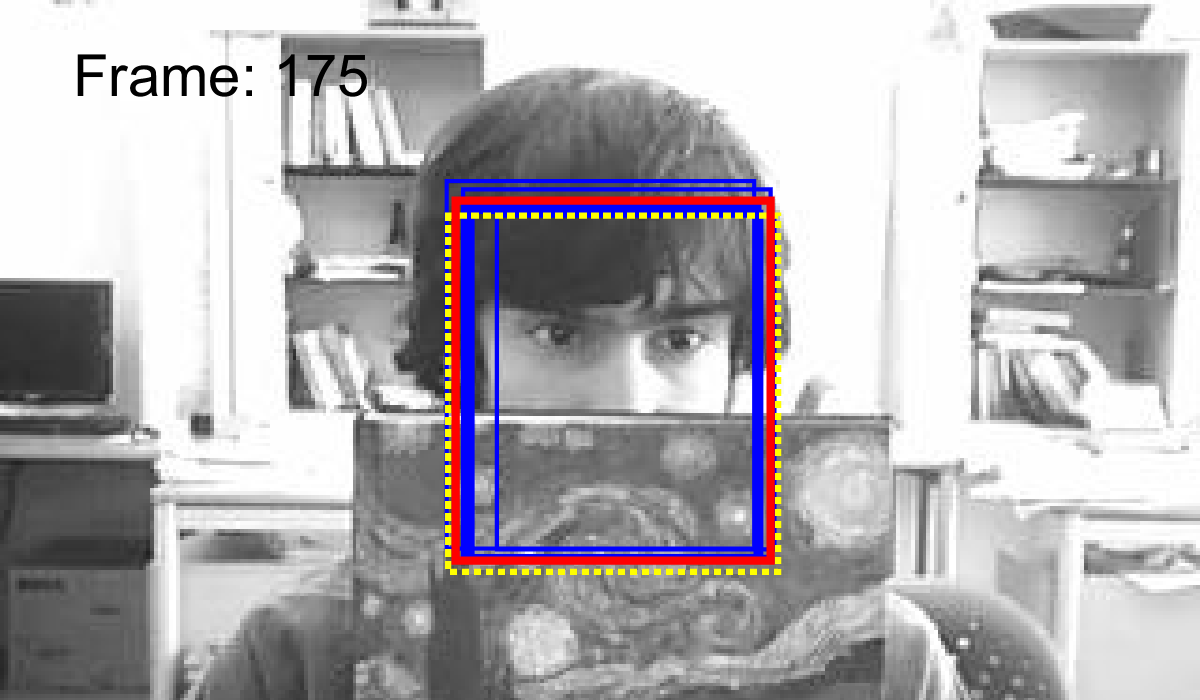}
\includegraphics[width= 0.32\linewidth]{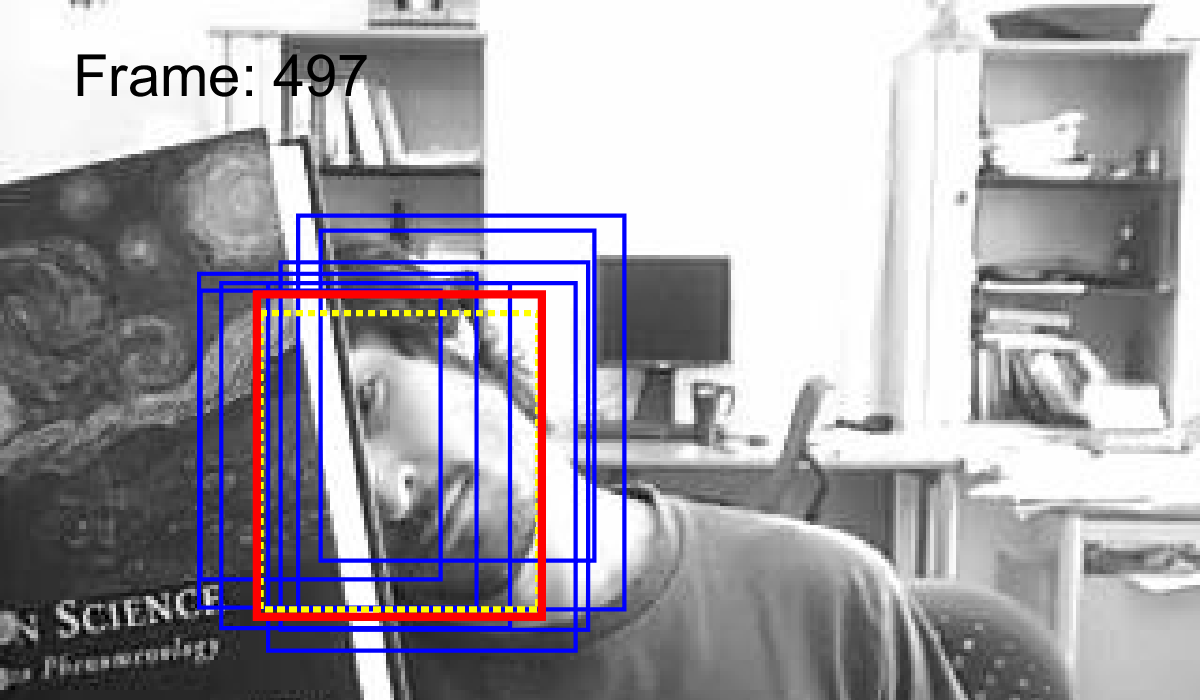}
\includegraphics[width= 0.32\linewidth]{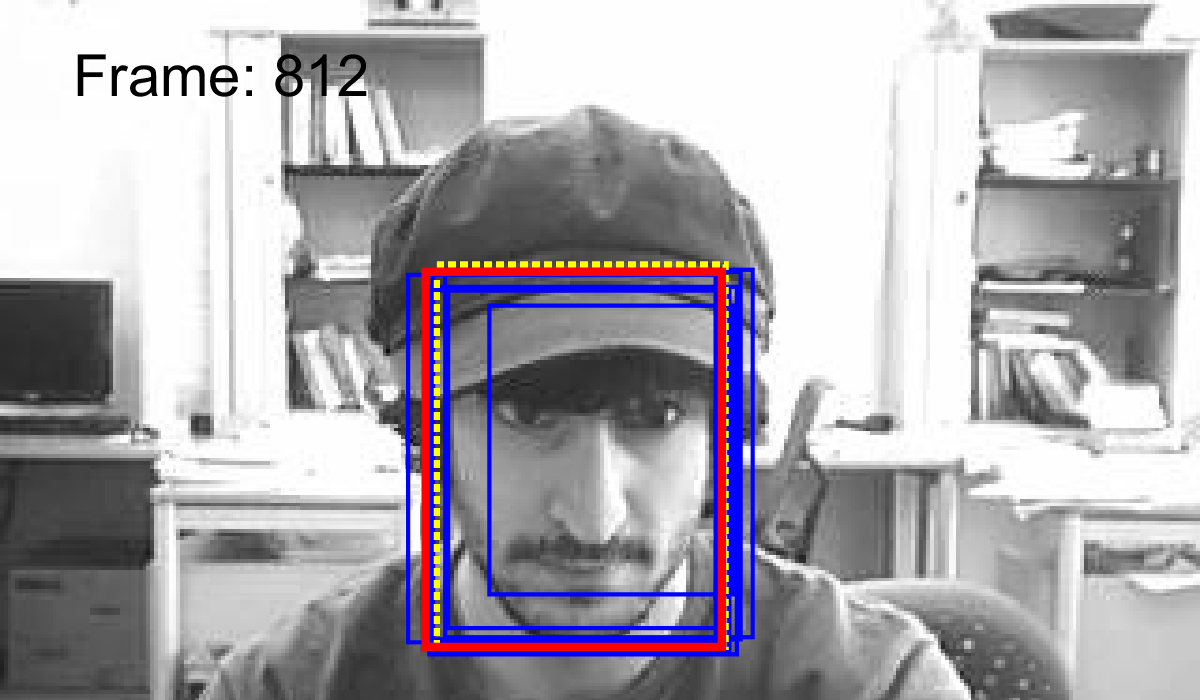}
\\
\includegraphics[width= 0.32\linewidth]{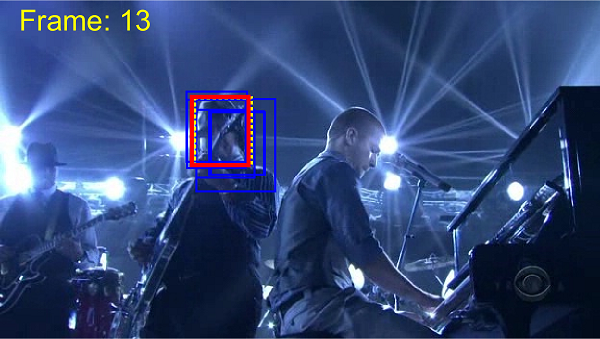}
\includegraphics[width= 0.32\linewidth]{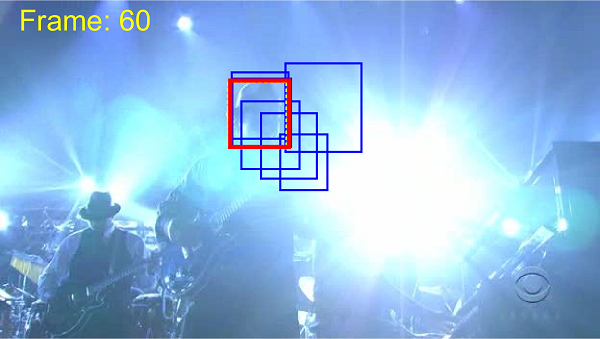}
\includegraphics[width= 0.32\linewidth]{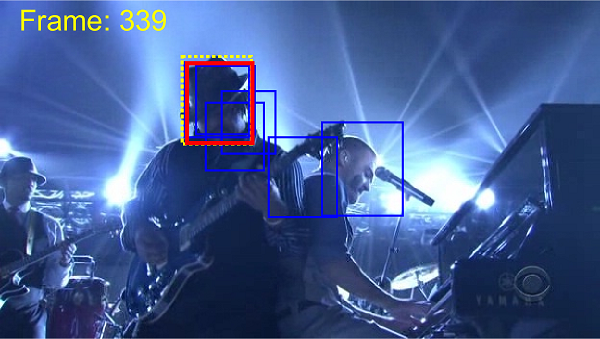}
\\
\includegraphics[width= 0.32\linewidth]{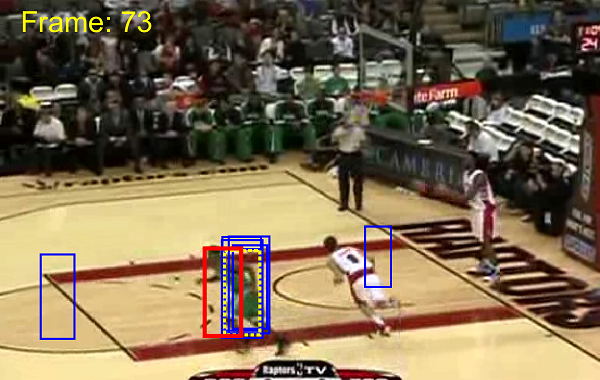}
\includegraphics[width= 0.32\linewidth]{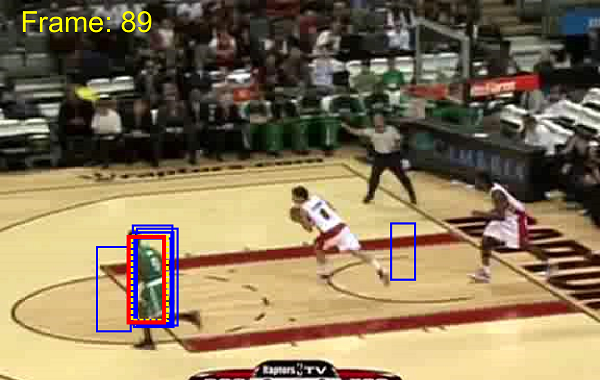}
\includegraphics[width= 0.32\linewidth]{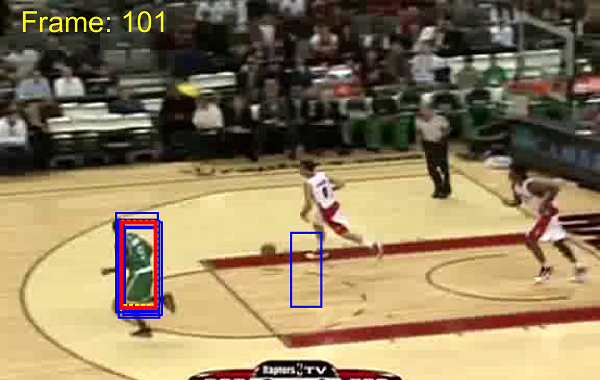}
\\
\includegraphics[width= 0.32\linewidth]{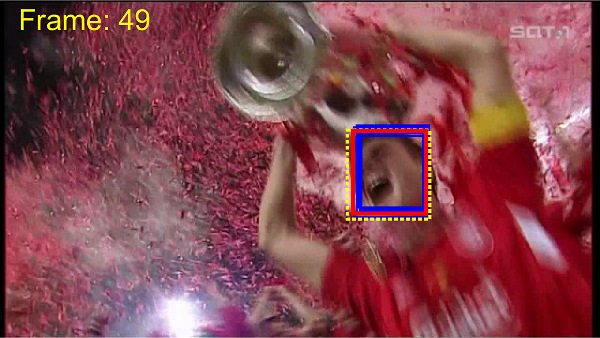}
\includegraphics[width= 0.32\linewidth]{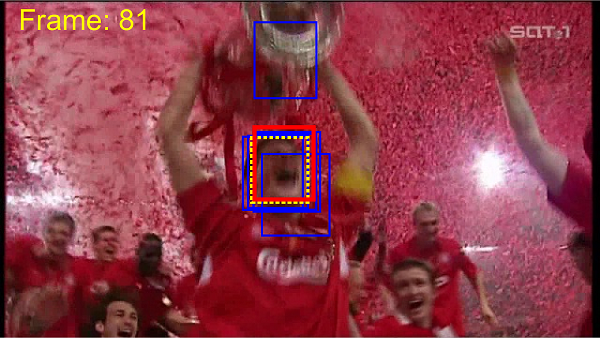}
\includegraphics[width= 0.32\linewidth]{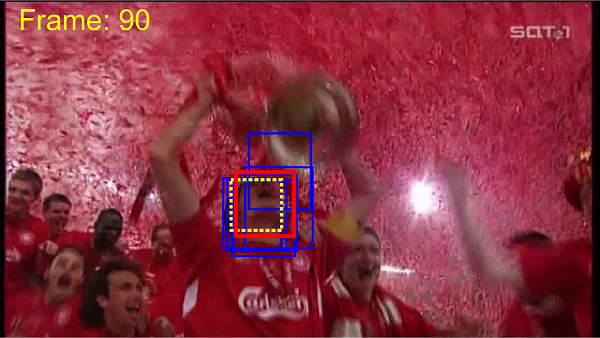}

\end{center}
\caption{Sample tracking results of evaluated algorithms on several challenging video sequences, in these sequences the red box depicts the DEDT against other trackers (blue). The ground truth is illustrated with yellow dashed box. From top to bottom the sequences are \textit{Skating1}, \textit{FaceOcc2}, \textit{Shaking}, \textit{Basketball}, and \textit{Soccer} with drastic illumination changes, scaling and out-of-plane rotations, background clutter, noise and severe occlusions.
}
\label{fig:qualitative}
\vspace{-1.2em}
\end{figure}

Figure \ref{fig:success} presents the success and precision plots of DEDT along with other state-of-the-art trackers for all sequences. It is shown in this plot that DEDT usually keeps the localization error under 10 pixels. Table \ref{tab:attributes} presents the area under the curve of the success plot (eq\eqref{eq:auc}) for all the sequences and their subcategories, each focusing on a certain challenge of the visual tracking. As shown, DEDT has the competitive precision compared to CCOT which employs state-of-the-art multi-resolution deep feature maps, and performs better than the rest of the other investigated trackers on this dataset. The performance of DEDT is comparable with CCOT in the case of illumination variation, deformation, out-of-view, out-of-plane rotation and motion blur, while it has superior performance in handling background clutter. This indicates the effectiveness of the target vs. background detection and flexibility for accommodating rapid target changes. While the former can be attributed to effective ensemble tracking, the latter is known to be the effect of combining long and short-term memory. 
%Given the strong detection characteristics of CCOT, derived from its features, it's superior performance in detecting occlusions and scale variations is reasonable, and a competitive performance of DEDT is expected with such features (in the cost of speed, however). 
It is observed in the run-time that for handling extreme rotations, the ensemble heavily relies on the auxiliary tracker, which although brings the superior performance in the category, a better representation of the ensemble model may reduce the reliance of the tracker to the auxiliary tracker.%(e.g. rotated bounding box) may improve the classification performance of the committee members and reduce their reliance on the auxiliary classifier. 
The proposed algorithm shows a sub-optimal performance in low-resolution scenario compared to DCF-based trackers (SRDCF, and CCOT), and although it does not provide a high-quality localization for smaller/low-resolution targets, it is able to keep tracking them. This finding highlights the importance of further research on the ensemble-based DCF trackers. 
Our method also achieved the best accuracy (0.58) on VOT2015 by outperforming SRDCF, yet the highest robustness (0.82) belongs to CCOT (Table \ref{tab:vot}).
Finally, a qualitative comparison of DEDT versus other trackers is presented in Figure \ref{fig:qualitative}.

\section{Conclusion}
In this study, we proposed diverse ensemble discriminative tracker (DEDT) that maintains a diverse committee of classifiers to the label of the samples and queries the most disputed labels --which are the most informative ones-- from a long-term memory auxiliary classifier. By generating artificial data with diverse labels, we intended to diversify the ensemble of classifiers, efficiently covering the version space, increasing the generalization of the ensemble, and as a result, improve the accuracy. In addition, by using the query-by-committee concept in labeling and updating stages of the tracker, the label noise problem is decreased.
By using the diverse committee, in turn, the problem of equal weights for the samples are addressed, and a good approximation of the target location is acquired even without dense sampling.  
The active learning scheme manages the balance between short-term and long-term memory by recalling the label from long-term memory when the short-term memory is not clear about the label (due to forgetting the label or insufficient data). This also reduces the dependence of the tracker on a single classifier (i.e., auxiliary classifier), yet breaking the self-learning loop to avoid accumulative model drift.  
The results of the experiment on OTB50, OTB100, and VOT2015 benchmarks demonstrate the competitive tracking performance of the proposed tracker compared with the state-of-the-art. %In the next step, we will investigate the strategies to detect or generate more challenging samples for the ensemble (e.g. adversarial samples of the ensemble) to accelerate the model construction especially in rapidly changing scenarios.   

\section*{Acknowledgment}
This study is partly supported by the Japan NEDO and the ``Post-K application development for exploratory challenges'' project of the Japan MEXT.

\clearpage
\appendix
\section{Comparison with Existing Studies}
From one hand, CMT\cite{meshgi2017active} uses multiple-memory horizons for obtaining training data and QBT\cite{meshgi2016robust} uses a simple bagging of most recent ``uncertain" data to update the classifier, but we construct artificial ``diversity" data from the distribution of most recent samples.

On the other hand, MUSTer\cite{hong2015multi} uses a long-term key-point database to validate whereas TGPR uses long-term memory to regularize the result of the short-memory tracker. Both use fixed heuristics to override the overall result of the short-memory tracker, after the tracking. For clarification, the novelties of the study are: building ensemble on disputed data and maintain it by the online update, diversify ensemble members by generating plausible artificial data, and active switch between short-long memories to label samples, where the short-long memory fusion is performed during labeling, and the data is exchanged between two memories.

\section{Elaboration on Proposed Idea}
Methods such as ensemble tracking are well-known for label noise and breaking self-learning loop. In addition, co-tracking framework \cite{tang2007co} breaks the self-learning loop by exchanging data between two parallel classifiers. We used both in a hierarchical fashion, and also utilized bagging as a part of ensemble model update, that promotes robustness against label noise. Furthermore, the batch update of the aux. classifier prevents the drift with a low amount of label noise and assists the ensemble to fight label noise in-turn.

On the other hand, since we used Gaussian sampling around the last target location, some samples (depending on the target type) are labeled positive, from which only the most similar one forms the tracker output, but others are used for retraining. For the initial frame, we perturbed the initial user bounding box of the target to generate initial training for both the ensemble and the aux. classifier.

The proposed diversification in each frame $t$, provides the ensemble member $\theta_t^{(c)}$ with a subset of training data. This makes the temporary ensemble $\mathcal{C}'_t$, trained on the obtained samples. Then the artificial data is from the sample's empirical distribution, but its label is selected in a way to challenge the ensemble's belief about those data. Once the model $\theta_t^{'(c)}$ is updated with generated ``diversity" samples, the total accuracy of the ensemble on all current samples is measured. If the accuracy improved, the ``diversity" samples are accepted, otherwise, new artificial samples are generated and the process repeats. By generating artificial data, the number of positive samples increases (samples are often negative $\rightarrow$ artificial data is often labeled positive), and since they are sampled from the data distribution (modeled by multivariate Gaussian), they are unlikely to be outliers.

\section{Combining Long and Short Memories}
\begin{figure}[b]
  \begin{center}
  \includegraphics[width=1\linewidth]{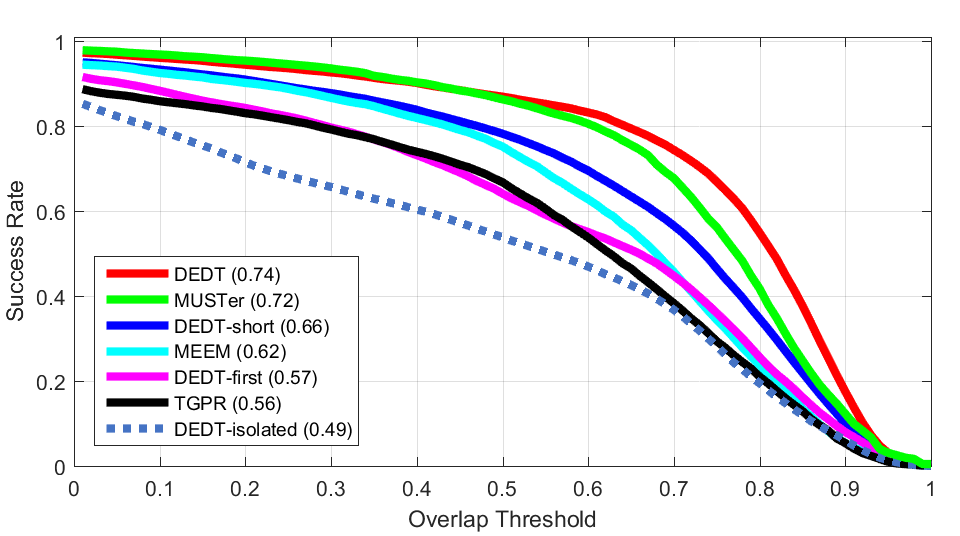} 
   \end{center}
  \caption{The effect of using long-term memory for auxiliary classifier of DEDT on the overall tracking results on OTB50 \cite{wu2013online}.}
\label{fig:long_term}
\end{figure}
Researchers have been combining long-term and short-term classifiers to realize robust tracking. TGPR categorized samples into auxiliary samples from early frames and update them slowly and carefully, and target samples from recent samples that are updated quickly and aggressively \cite{gao2014transfer}. MEEM selects an snapshot of the classifier trained by the samples obtained from the beginning of the tracking to role-back inappropriate updates of the classifier \cite{zhang2014meem}. MUSTer archives consistent key-points of the target in he long-term memory, and validates the tracking of the short-term tracker \cite{hong2015multi}. In our proposed tracker, however, an ensemble of short-memory classifiers invoke the long-term memory when deemed necessary and an active query mechanism governs this process to balance the use of long and short term memories. To see the performance of this scheme, we made DEDT-first that trains the auxiliary classifier on the first frame and do not update this classifier, DEDT-short that updates the auxiliary classifier on each frame, canceling its long-memory properties, and DEDT-isolated that isolate the ensemble from auxiliary classifier, and fuse their results in the end similar to \cite{hong2015multi}. Figure \ref{fig:long_term} shows that both of such strategies have inferior performance in our settings, which promotes the role of active query selection and loopy update of the auxiliary classifier.

\begin{figure*}
\centering
\subfigure[ALL (CCOT, DEDT, CMT)\label{fig:all}]{\includegraphics[width= 0.32\linewidth]{success_plot_ALL_OTB50}}
\subfigure[IV (DEDT, CCOT, CMT)\label{fig:iv}]{\includegraphics[width= 0.32\linewidth]{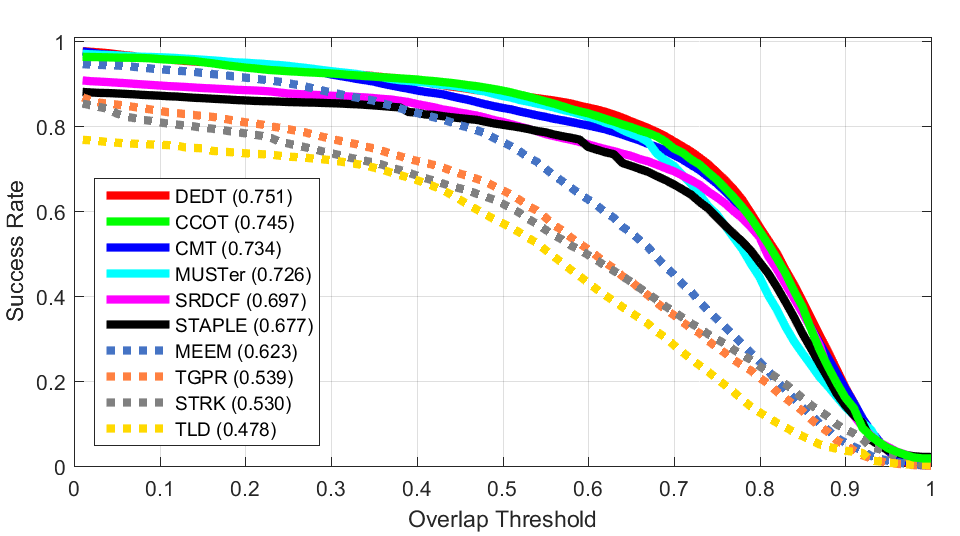}}
\subfigure[SV (CCOT, DDET, CMT)\label{fig:sv}]{\includegraphics[width= 0.32\linewidth]{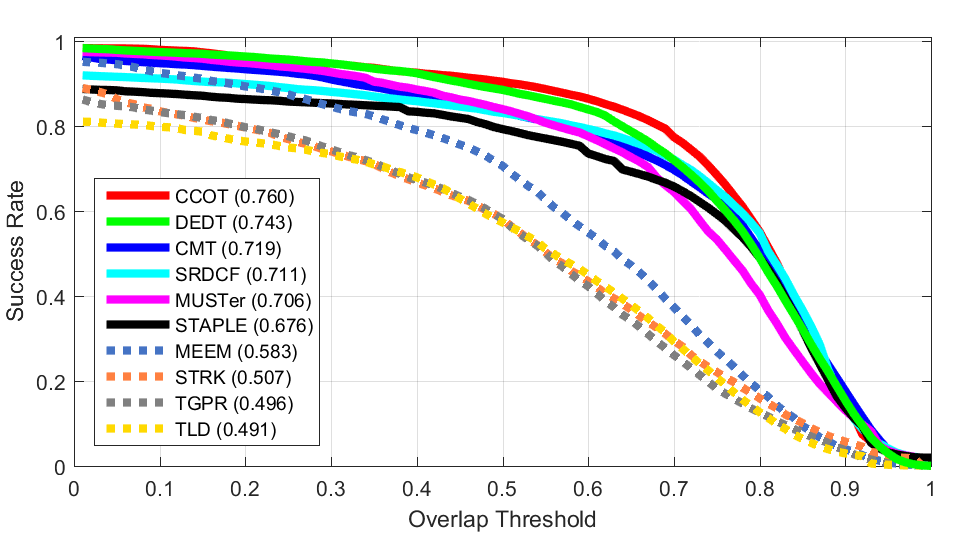}}
\subfigure[OCC (CCOT, DDET, MUSTer)\label{fig:occ}]{\includegraphics[width= 0.32\linewidth]{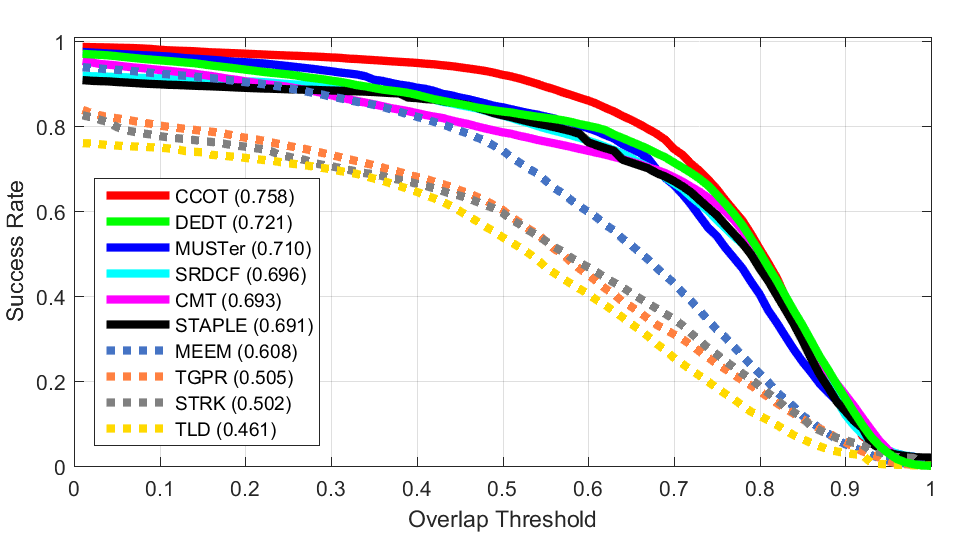}}
\subfigure[DEF (STAPLE, DDET, CMT)\label{fig:def}]{\includegraphics[width= 0.32\linewidth]{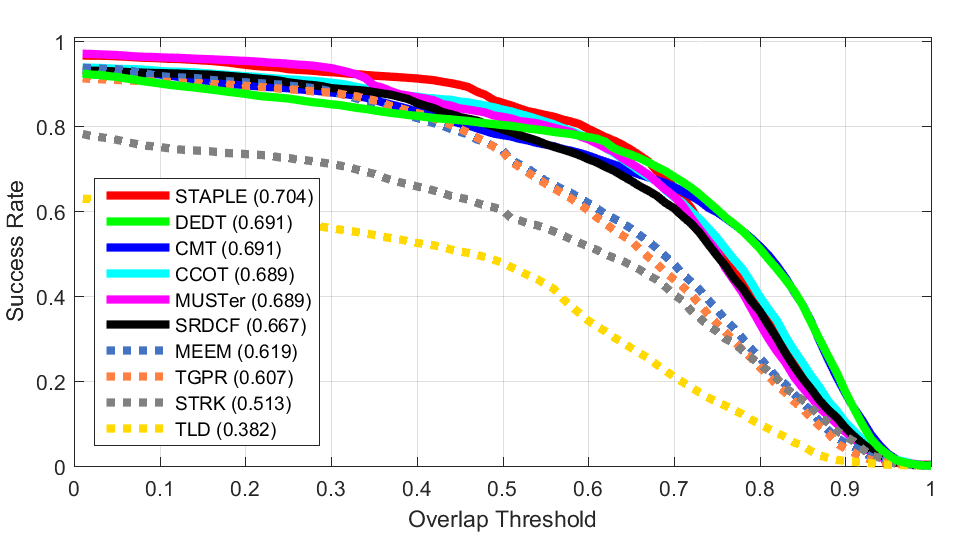}}
\subfigure[IPR (CMT, DEDT, CCOT)\label{fig:ipr}]{\includegraphics[width= 0.32\linewidth]{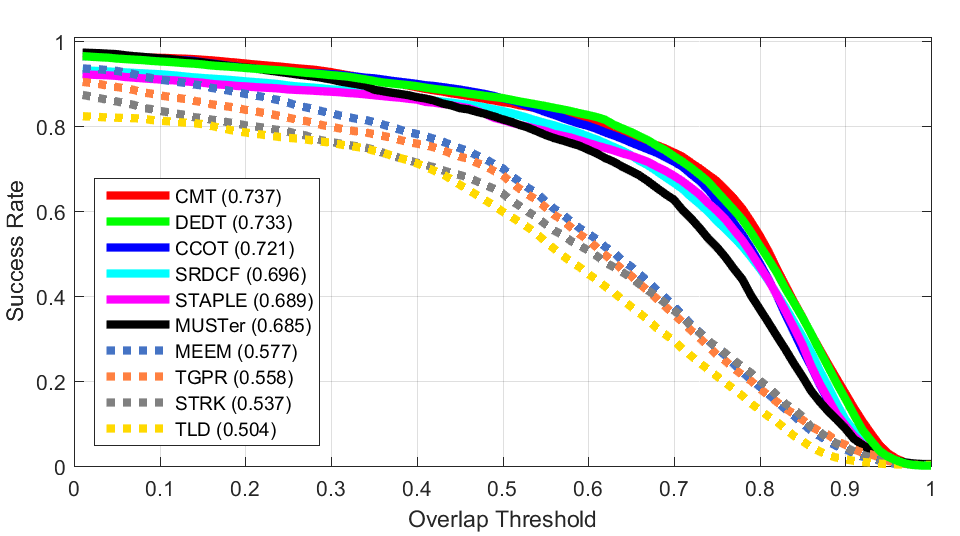}}
\subfigure[OPR (DEDT, CCOT, VTS)\label{fig:opr}]{\includegraphics[width= 0.32\linewidth]{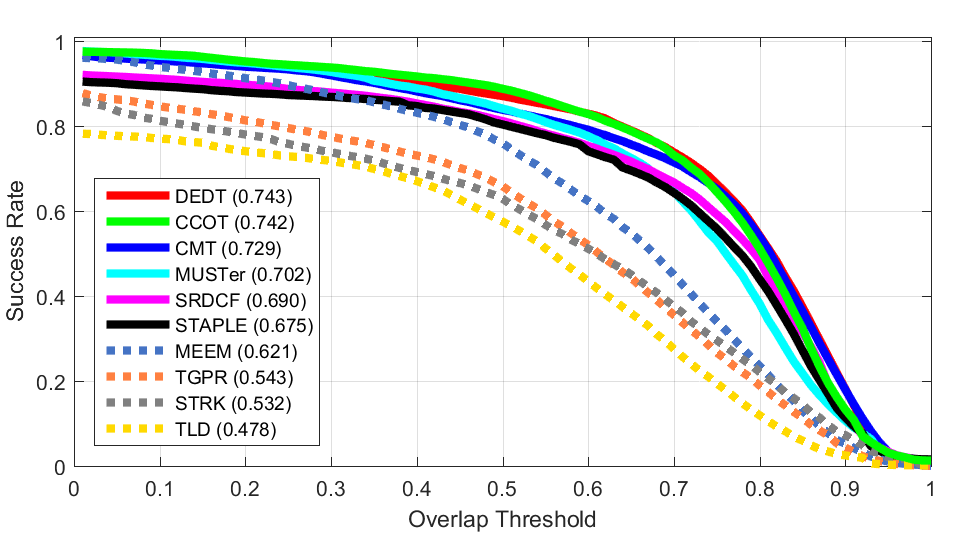}}
\subfigure[OV (CCOT, DDET, MUSTer)\label{fig:ov}]{\includegraphics[width= 0.32\linewidth]{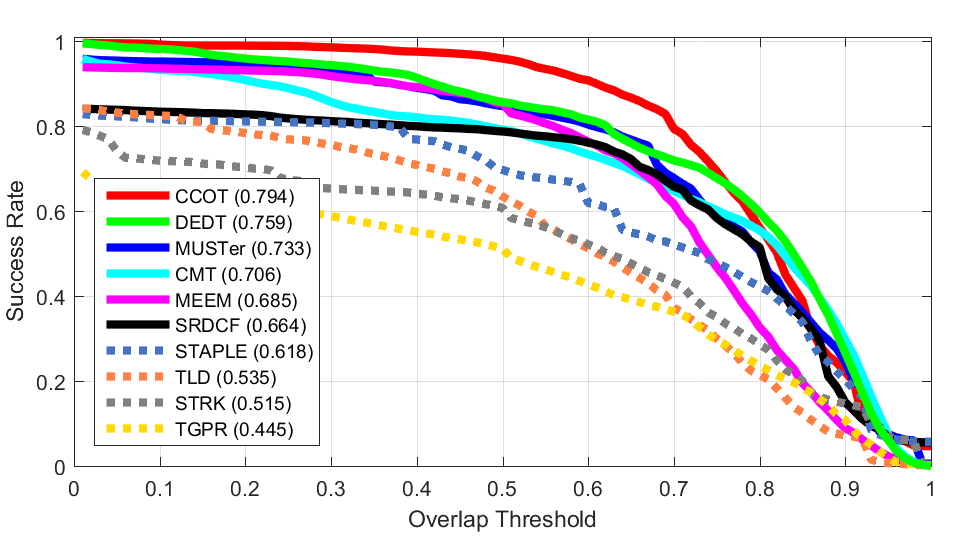}}
\subfigure[BC (DEDT, MUSTer, CCOT)\label{fig:bc}]{\includegraphics[width= 0.32\linewidth]{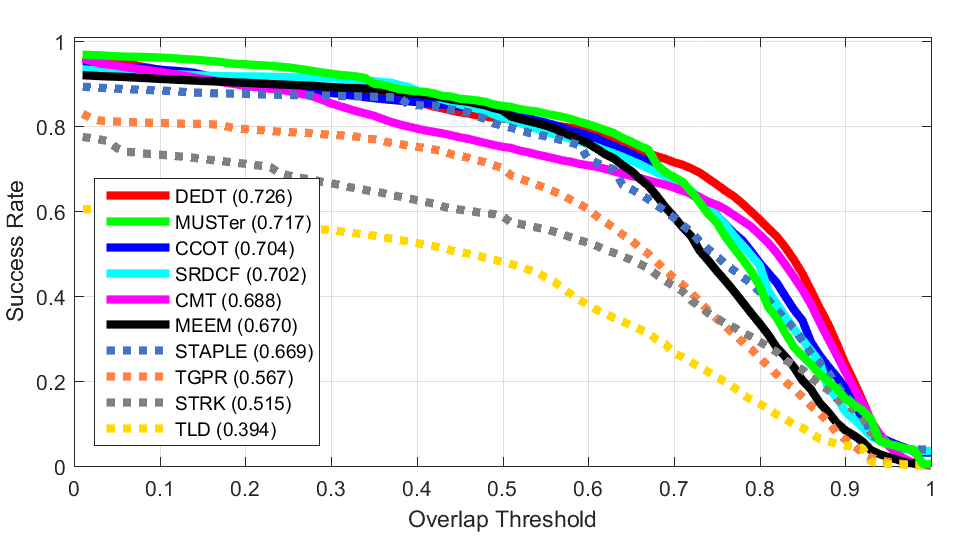}}
\subfigure[LR (CCOT, SRDCF, DEDT)\label{fig:lr}]{\includegraphics[width= 0.32\linewidth]{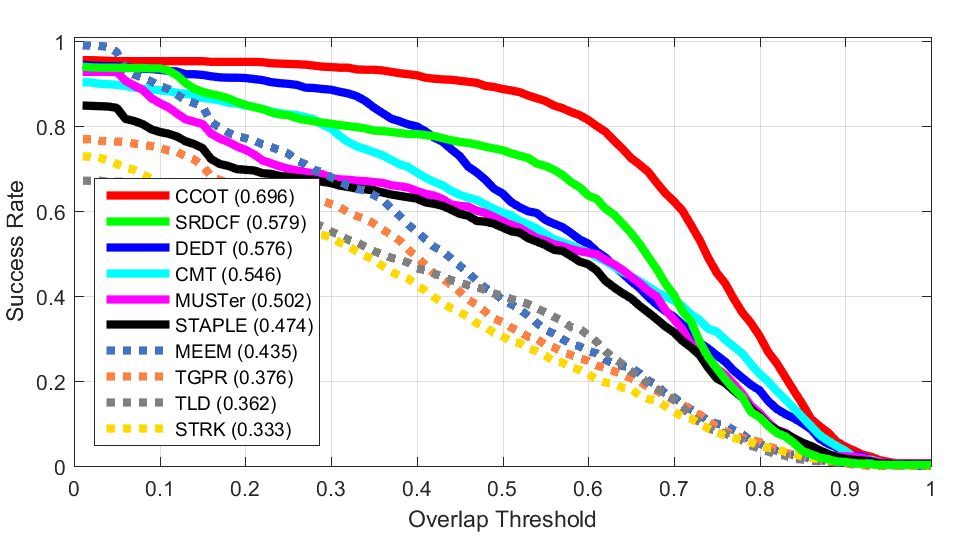}}
\subfigure[FM (DEDT, CCOT, CMT)\label{fig:fm}]{\includegraphics[width= 0.32\linewidth]{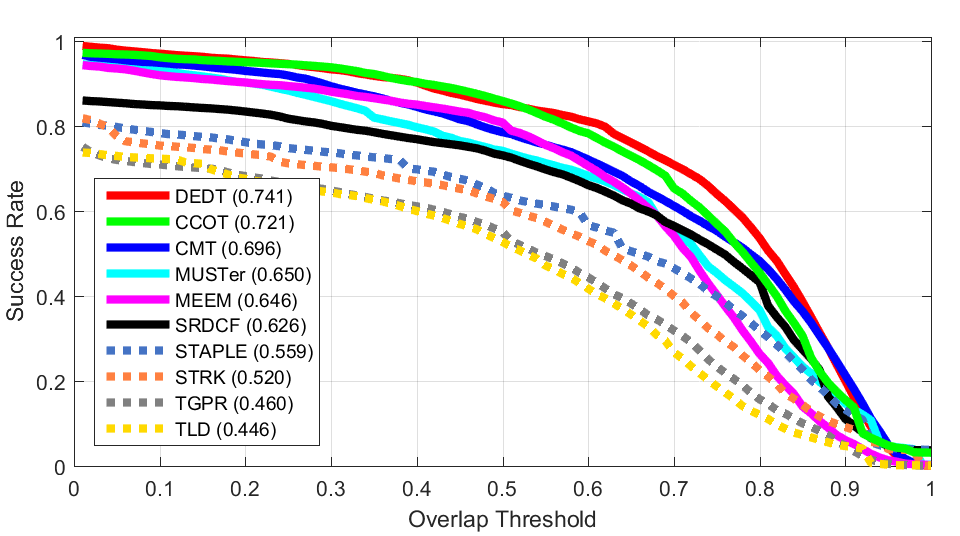}}
\subfigure[MB (CCOT, DEDT, SRDCF)\label{fig:mb}]{\includegraphics[width= 0.32\linewidth]{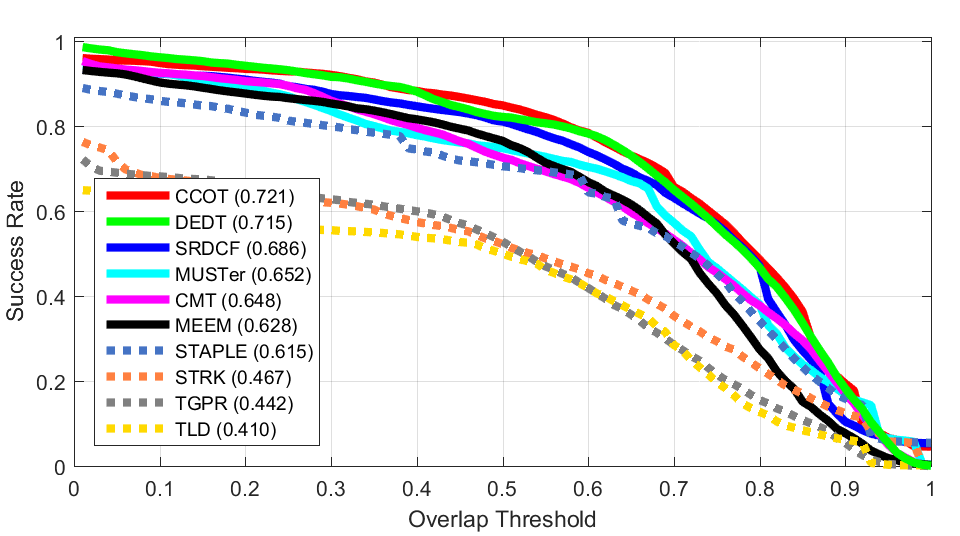}}
\caption{Quantitative evaluation of trackers under different visual tracking challenges (Top three performing trackers are listed in the order of their $AUC$ values). The {\color{red}DEDT} is plotted against other state-of-the-art algorithms. DEDT outperformed other trackers (except in overall and DEF (Fig. \ref{fig:def}) category) when dealing with different tracking challenges of OTB50 \cite{wu2013online} at all of the subcategories. It is shown in \ref{fig:all} that DEDT, clearly has a better overall performance compared to other trackers.}
\label{fig:eval_succ_all}
\vspace{-0.5 cm}
\end{figure*}

\section{Discussion}
\begin{figure*}[!t]
\centering
\subfigure[ALL (CCOT, DEDT, SRDCF)\label{fig:all100}]{\includegraphics[width= 0.32\linewidth]{success_plot_ALL_OTB100}}
\subfigure[IV (CCOT, DEDT, SRDCF)\label{fig:iv100}]{\includegraphics[width= 0.32\linewidth]{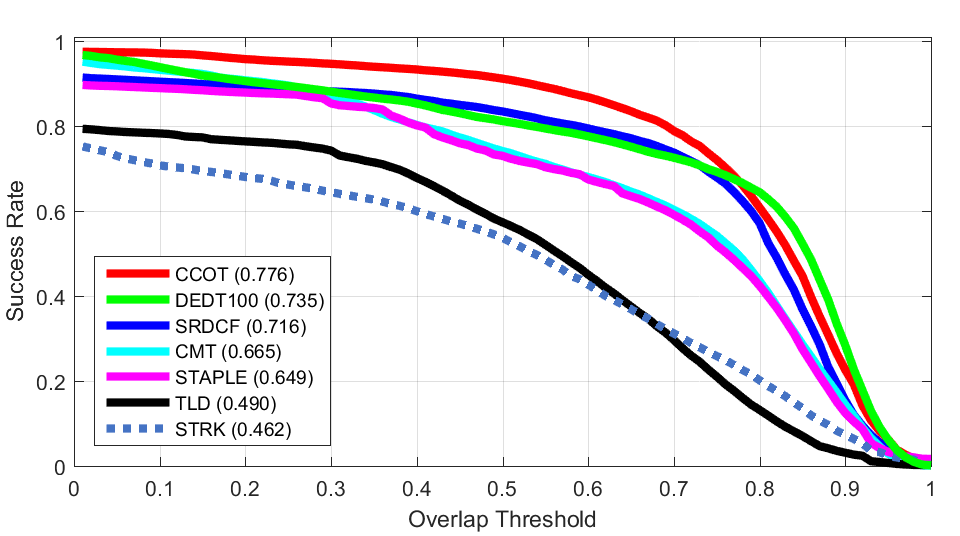}}
\subfigure[SV (CCOT, DEDT, SRDCF)\label{fig:sv100}]{\includegraphics[width= 0.32\linewidth]{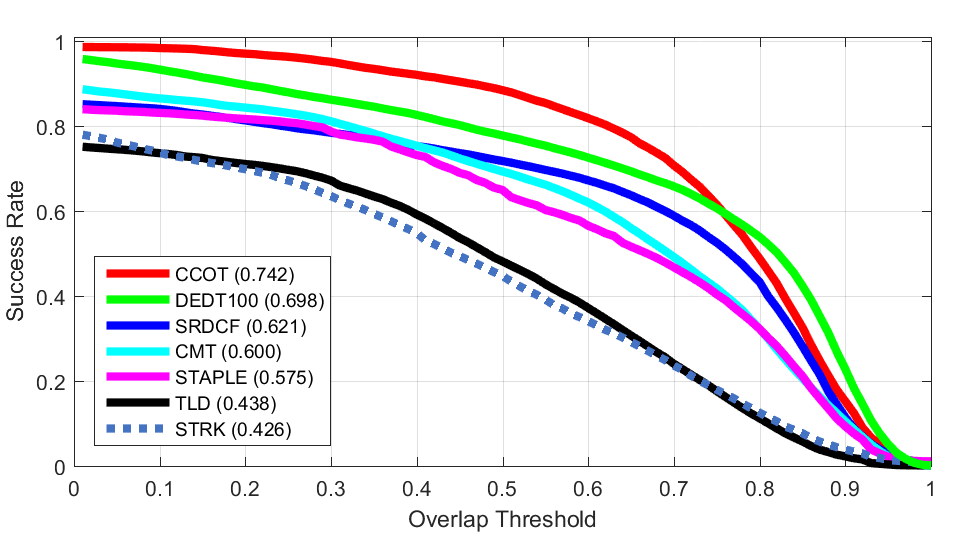}}
\subfigure[OCC (CCOT, DEDT, STAPLE)\label{fig:occ100}]{\includegraphics[width= 0.32\linewidth]{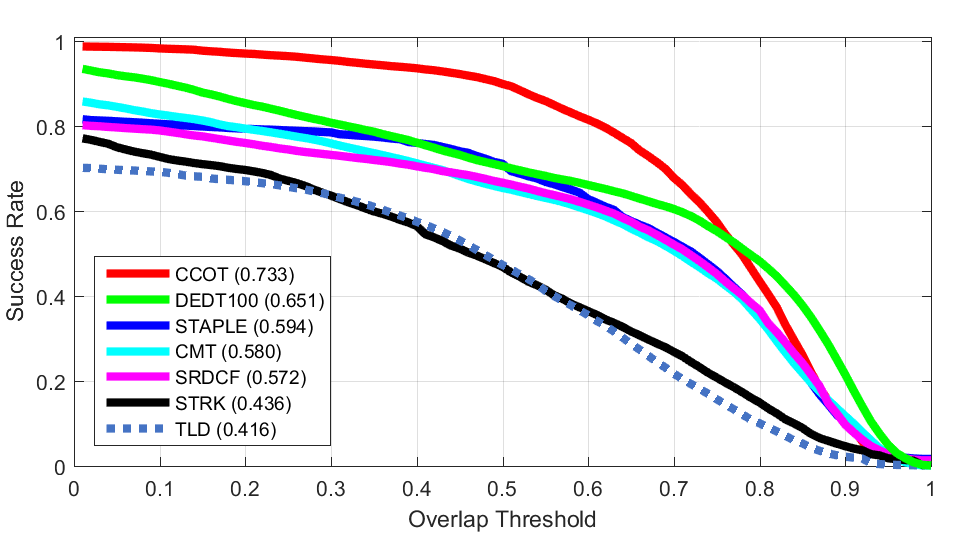}}
\subfigure[DEF (CCOT, DEDT, STAPLE)\label{fig:def100}]{\includegraphics[width= 0.32\linewidth]{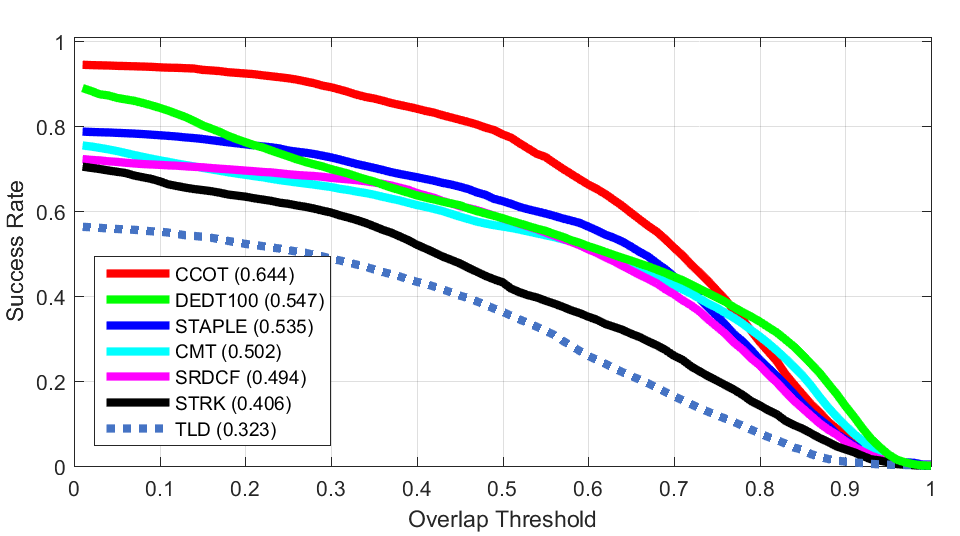}}
\subfigure[IPR (CCOT, DEDT, CMT)\label{fig:ipr100}]{\includegraphics[width= 0.32\linewidth]{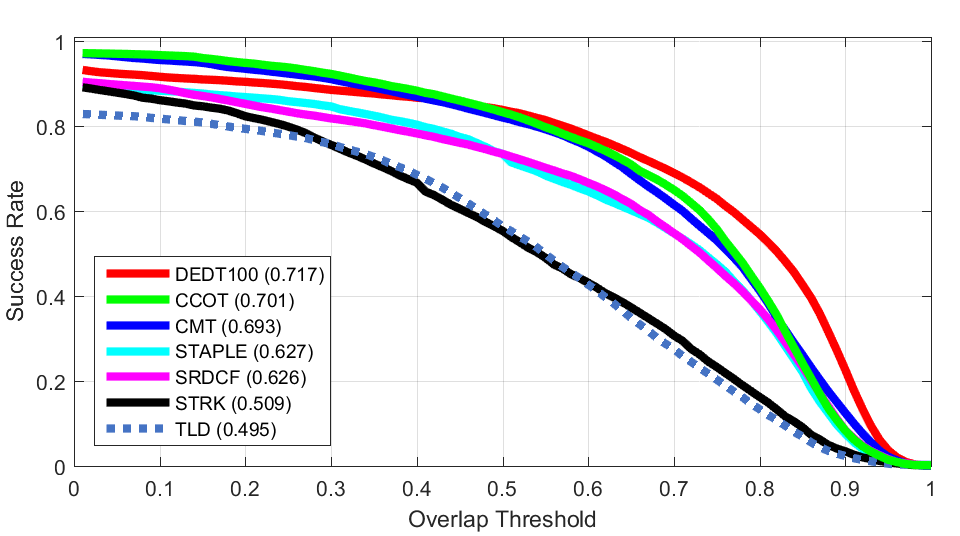}}
\subfigure[OPR (CCOT, DEDT, CMT)\label{fig:opr100}]{\includegraphics[width= 0.32\linewidth]{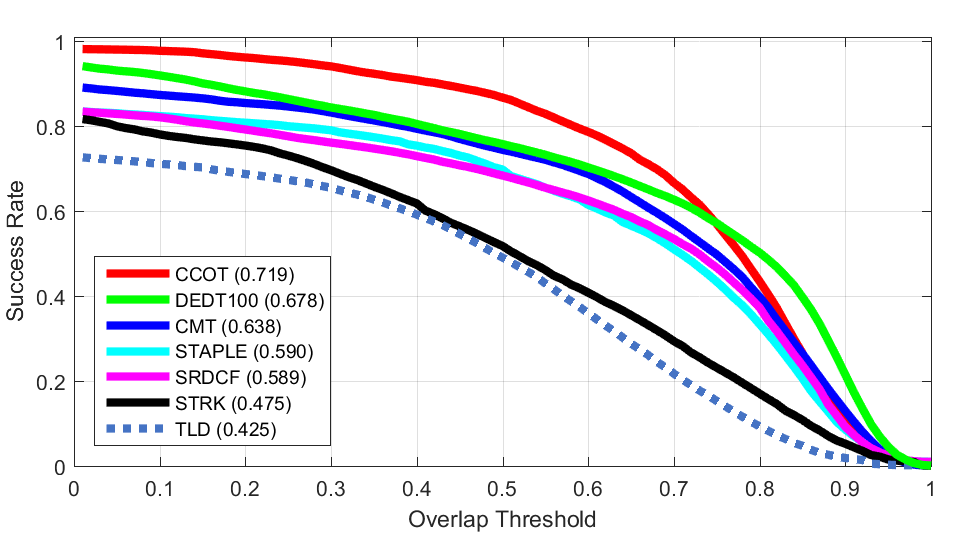}}
\subfigure[OV (CCOT, DEDT, CMT)\label{fig:ov100}]{\includegraphics[width= 0.32\linewidth]{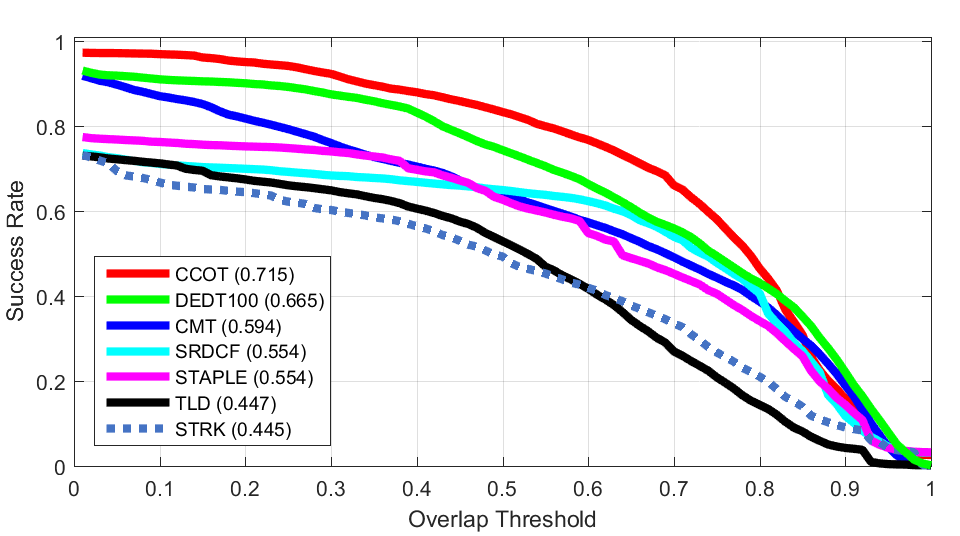}}
\subfigure[BC (CCOT, DEDT, SRDCF)\label{fig:bc100}]{\includegraphics[width= 0.32\linewidth]{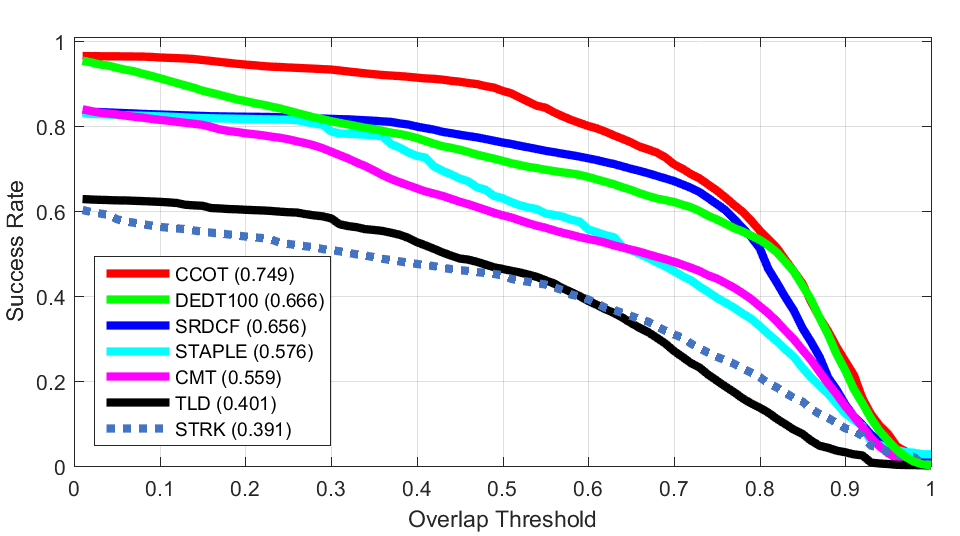}}
\subfigure[LR (CCOT, SRDCF, DEDT)\label{fig:lr100}]{\includegraphics[width= 0.32\linewidth]{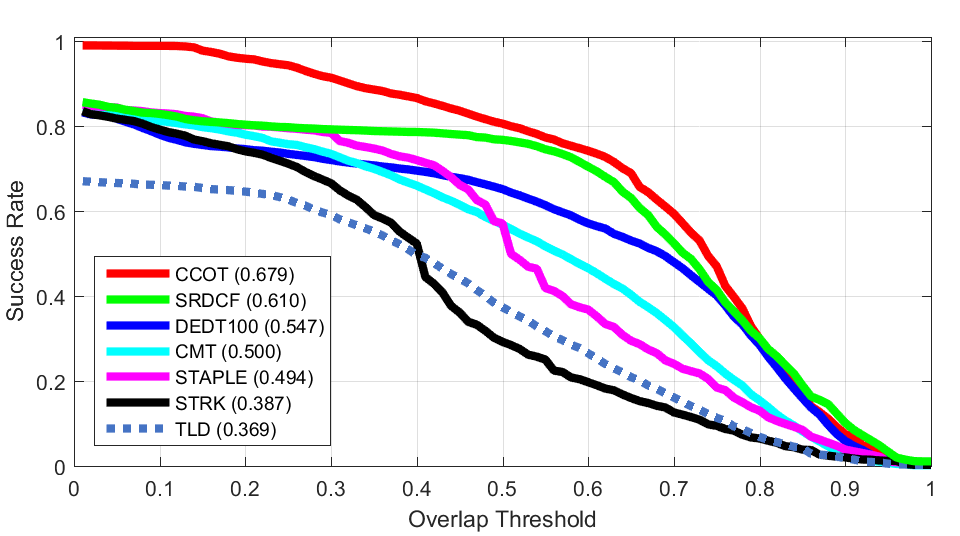}}
\subfigure[FM (CCOT, DEDT, CMT)\label{fig:fm100}]{\includegraphics[width= 0.32\linewidth]{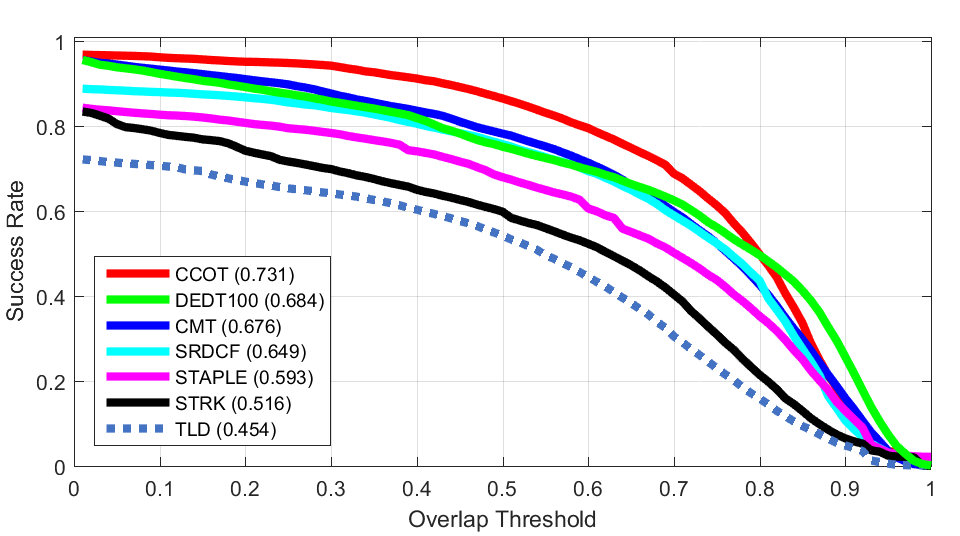}}
\subfigure[MB (CCOT, DEDT, SRDCF)\label{fig:mb100}]{\includegraphics[width= 0.32\linewidth]{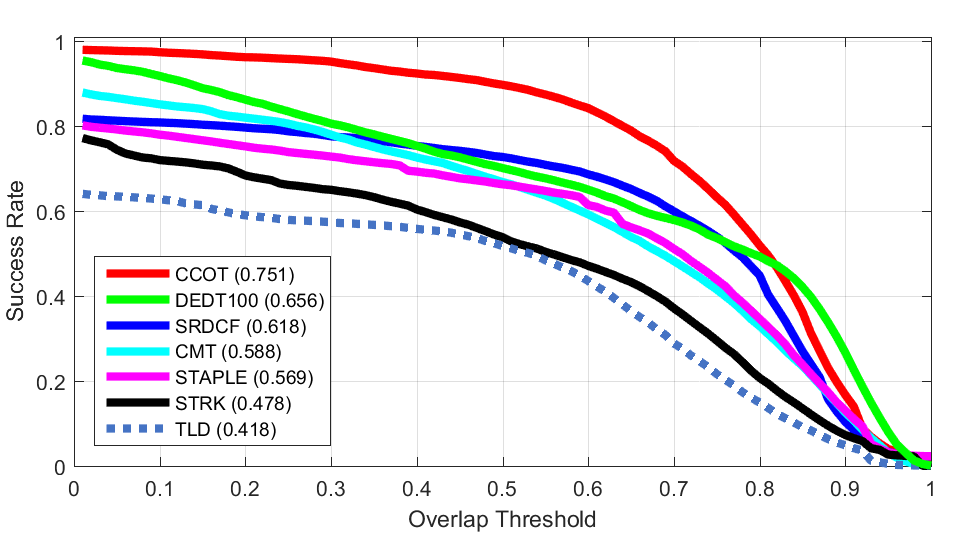}}
\caption{Quantitative evaluation of trackers under different visual tracking challenges (Top three performing trackers are listed in the order of their $AUC$ values). The {\color{red}DEDT} is plotted against other state-of-the-art algorithms. Except CCOT, DEDT outperformed other trackers (except in the LR \ref{fig:lr100} category) when dealing with different tracking challenges of OTB100 \cite{wu2015object} at all of the subcategories. It is shown in \ref{fig:all} that CCOT and DEDT, clearly has an edge comparing to other trackers, while CCOT employs deep features and DEDT uses HOG.}
\label{fig:eval_succ_all100}
\vspace{-0.5 cm}
\end{figure*}

The proposed tracker tackled some of the important topics in tracking community: noisy labels, sparse positive samples, and model drift due to self-learning loop. 

To alleviate label noise and breaking self-learning loop, methods such as ensemble tracking has been established in the literature. In addition, co-tracking framework \cite{tang2007co} breaks the self-learning loop by exchanging data between two parallel classifiers. We used both of them in a hierarchical fashion, and also we utilized bagging as a part of ensemble model update, that promotes robustness against label noise. Furthermore, the batch update of the auxiliary classifier prevents the drift with a small amount of label noise and serves as the helper of the ensemble to fight label noise in-turn.

On the other hand, since we used Gaussian sampling around the last target location, some of the samples (depending on the target type) are labeled positive, from which only the most similar one is considered as the tracker output, but the others are used for positive samples in the retraining. For the initial frame, following a popular routine, we perturbed the initial user-annotated bounding box of the target to generate initial training for both the ensemble and the aux. classifier.

The proposed diversification mechanism, in each frame $t$, provides the ensemble member $\theta_t^{(c)}$ with a subset of the training data (which we ensured to have enough positive data in the implementation). This makes the temporary ensemble $\mathcal{C}'_t$, trained on the obtained samples. Then the artificial data is generated using the same distribution of the samples, but its label is selected in a way to challenge the belief of the ensemble about such data. Once the model $\theta_t^{'(c)}$ is updated with this generated "diversity" samples, the total accuracy of the ensemble on all current samples is measured. If the accuracy was improved, the "diversity" samples are accepted, otherwise, new artificial samples are generated and the same routine repeats. By generating artificial data, the number of positive samples increases (samples are usually negative $\rightarrow$ artificial data is usually labeled positive), and since they are sampled from the data distribution (modeled by multivariate Gaussian here), these samples are unlikely to be outliers.

Detailed success plots of comparisons against state-of-the-art trackers on OTB50 and OTB100 datasets are provided in Figures \ref{fig:eval_succ_all} and \ref{fig:eval_succ_all100} respectively. 

{\small
\balance %%%%%%%%%%%%%%%%%% to do
\bibliographystyle{ieee}
\bibliography{refs}
}

\end{document}